\documentclass[11pt]{article}

\usepackage[preprint]{acl}

\usepackage{times}
\usepackage{latexsym}

\usepackage[T1]{fontenc}

\usepackage[utf8]{inputenc}

\usepackage{microtype}

\usepackage{inconsolata}

\usepackage{graphicx}
\usepackage{algorithm}
\usepackage{algorithmic}
\usepackage{booktabs}

\usepackage{multirow}
\usepackage{booktabs}
\usepackage{amsmath,amssymb}
\usepackage{siunitx}
\usepackage{soul} 
\usepackage{xcolor}
\usepackage{tabularx}
\usepackage{subcaption}

\title{Asymmetric Capacity Allocation in Self-Refinement Pipelines}

\author{
  Zhuoyi Yang$^{1}$ \quad
  Ian G. Harris$^{1}$ \quad
  Salar Hashemitaheri$^{1}$ \quad
  Cassie Huang$^{2}$ \quad
  Yuangang Li$^{1}$ \\
  Hyunwoo Oh$^{1}$ \quad
  Paul Dourish$^{1}$ \quad
  Tony Givargis$^{1}$ \quad
  Mohsen Imani$^{1}$ \quad
  Li Zhang$^{2}$ \\[1ex]
  $^{1}$University of California, Irvine \\
  $^{2}$Drexel University \\
  \texttt{
    \{zhuoyy1, iharris, salarh, yuanganl, hyunwooo, givargis, mohseni\}@uci.edu
  } \\
  \texttt{jpd@ics.uci.edu} \\
  \texttt{\{Cassie.Huang, Harry.Zhang\}@drexel.edu}
}

\begin{document}
\maketitle
\begin{abstract}

Self-refinement, typically structured as generation, critique, and revision, is a widely adopted paradigm for improving LLM generation and serves as a core mechanism in many LLM agents. While the three stages involve different cognitive demands, most existing approaches conveniently treat the model size as an implementation detail rather than a subject of study, which may lead to a waste of resources. Little work has systematically examined how model size affects each stage or whether effective self-refinement requires equally capable models for generation, critique, and revision. We present the first stage-wise model size study of the self-refinement pipeline on 5 benchmarks from different domains using 6 model sizes of Qwen3 and 4 model sizes of Gemma 3. We conclude that larger generators and refiners generally improve the pipeline, whereas an undersized refiner can even harm performance. Second, performance is highly insensitive to the size of the critic, although including even a small critic consistently outperforms omitting critique altogether. Our findings demonstrate that model capacity should not be allocated uniformly across self-refinement pipelines. Instead, different stages exhibit distinct size scaling characteristics, providing practical guidance for designing more computationally efficient multi-stage language model systems.
\end{abstract}

\begin{figure*}[t]
    \centering
    \includegraphics[width=\textwidth]{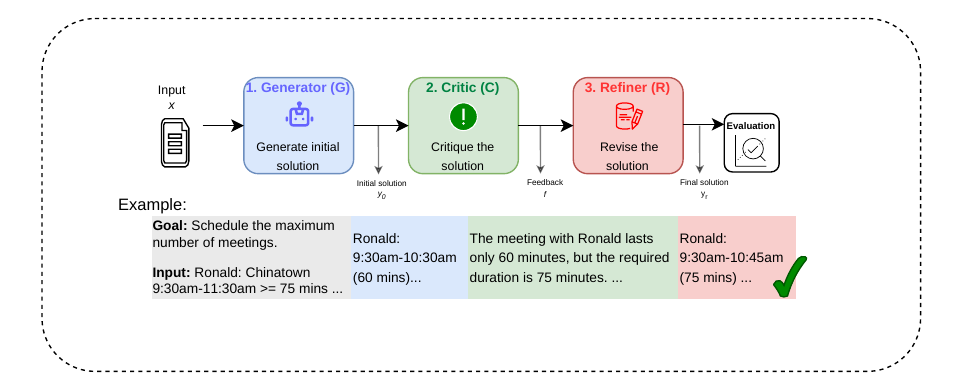}
    \caption{Overview of our three-stage self-refinement pipeline with a simplified example. Given an input $x$, the generator produces an initial solution $y_0$, the critic provides natural-language feedback $f$ on $x$ and $y_0$, and the refiner revises the solution based on $x$, $y_0$, and $f$.}
    \label{fig:pipeline}
\end{figure*}

\section{Introduction}
Large language models increasingly rely on iterative reasoning rather than one-shot generation. A common paradigm is self-refinement \citep{madaan2023selfrefine}, in which a model first produces a solution, critiques its own output, and then revises the solution based on the generated feedback. This generator--critic--refiner pipeline has become a fundamental component of modern LLM systems, improving performance on reasoning, planning, coding, and generation tasks \citep{shinn2023reflexion, CRITIC, refinecode, self-debug}. More broadly, similar refinement loops are now embedded within many agentic workflows, where agents repeatedly evaluate intermediate outputs before taking subsequent actions \citep{agentsurvey, autogen, react}.

 Existing work typically selects model sizes heuristically or uniformly scales the entire pipeline, without examining whether all stages benefit equally from increased model capacity \citep{madaan2023selfrefine, shinn2023reflexion, CRITIC}. Such design likely leads to a waste of resources. The decomposition of generator-critic-refiner is also consistent with classical cognitive models that distinguish production, evaluation, and revision as separate cognitive processes \citep{hayes1980identifying, hayes1996new, kellogg1996model}. The three stages perform fundamentally different functions but model capacities in the pipeline have not been treated as a subject of systematic investigation. This raises a fundamental question: How should model capacity be allocated across the stages of a self-refinement pipeline?

We address this question through a controlled stage-wise model size analysis on the self-refinement pipeline (Figure \ref{fig:pipeline}). We independently vary the size of the generator, critic, and refiner while holding the other two stages fixed (Figure \ref{fig:scaling_matrix}). This experimental design isolates the contribution of each stage and enables a direct comparison of their sensitivity to model sizes across 5 benchmarks and 2 model families (Qwen3 \citep{qwen3} and Gemma 3 \citep{gemma3}).
We draw two main conclusions. First, self-refinement performance is highly sensitive to the capacities of the generator and refiner. Scaling either stage generally improves overall performance, whereas an undersized refiner can even degrade performance below that of the initial generation. Second, pipeline performance is comparatively insensitive to critic size. Although larger critics provide only marginal gains, even the smallest critic consistently outperforms a pipeline without explicit critique. Our findings show that the three stages exhibit markedly different scaling characteristics, indicating that uniform model allocation is suboptimal. These results provide practical guidance for designing computationally efficient self-refinement pipelines.

\section{Related Works}

\textbf{Self-refinement} has become a widely adopted paradigm for improving large language models at inference time. Early work such as SELF-REFINE demonstrated that a single language model can iteratively generate feedback and refine its own outputs without additional training \citep{madaan2023selfrefine}. Since then, numerous approaches have extended this paradigm by incorporating stronger critics, external verification, execution feedback, retrieval, and tool use to improve reasoning, planning, coding, and text generation \citep{shinn2023reflexion, refinecode, iterativetranslation, self-debug}. More recently, the generate–critique–refine loop has become one of the core reasoning primitives underlying many contemporary agentic systems. \citep{react, llmcompiler, autogen}. Recent surveys likewise identify reflection and self-correction as central cognitive components of modern agent architectures \citep{agentsurvey, understandingplanningllmagents, generalizability}. Despite the diversity of refinement strategies, existing work primarily focuses on designing more effective feedback mechanisms or refinement algorithms to maximize downstream task performance, while typically treating the model assigned to each stage as fixed.


A separate line of research investigates how model capacity and computational budget influence language model performance. Classical scaling law studies characterize how performance improves with increasing model size, training data, and compute, providing principles for efficient model development \citep{kaplanscaling, chinchila}. More recently, researchers have shifted attention from training-time scaling to inference-time scaling, showing that allocating additional computation during reasoning can substantially improve performance without increasing model parameters. Representative approaches include deliberate reasoning through longer inference trajectories \citep{o1, deepseek_r1}, self-guided reasoning and internal thought generation \citep{quietstar}, and test-time compute scaling strategies that trade additional inference computation for higher accuracy \citep{s1, budget_forcing}. 

Complementary work studies how inference computation should be allocated, including LLM routing \citep{RouteLLM}, model cascading \citep{frugalGPT}, adaptive model selection \citep{BEST-route}, and mixture-of-experts routing \citep{Switch_transformers, Gshard}, which dynamically select models or experts according to input difficulty or computational budget. While these methods determine how much computation to spend on each query or which model should answer it, they generally treat inference as a single prediction stage. In contrast, we study how model capacity should be distributed across the interacting stages of a multi-stage self-refinement pipeline.

Although self-refinement pipelines and model size scaling have each been studied extensively, their intersection remains largely unexplored. To the best of our knowledge, no prior work has systematically investigated how model capacity should be allocated across the generator, critic, and refiner in a self-refinement pipeline. We address this gap by conducting the first systematic stage-wise model size analysis of self-refinement pipelines.

\section{Method}
This section describes our experimental framework for analyzing model capacity in self-refinement pipelines. We first formalize the generator--critic--refiner pipeline, then present our stage-wise model size analysis protocol.

\subsection{Self-Refinement Pipeline}


Our self-refinement framework consists of three sequential stages: a \emph{generator}, a \emph{critic}, and a \emph{refiner}, following the SELF-REFINE paradigm proposed by \citet{madaan2023selfrefine}.
Given an input $x$, the generator first produces an initial solution

\[
y_0 = G(x),
\]

where $G$ denotes the generator. The critic then analyzes the generated solution and produces natural-language critique

\[
f = C(x, y_0),
\]

where $C$ denotes the critic. $f$ includes potential errors and actionable critique. Finally, the refiner generates the final solution by applying the critique to the original solution,

\[
y_r = R(x, y_0, f),
\]

where $R$ denotes the refinement model.

We denote a complete self-refinement pipeline by

\[
P(G, C, R),
\]

where $G$, $C$, and $R$ specify the models used for the generator, critic, and refiner. For example,

\[
P(32\mathrm{B},\,0.6\mathrm{B},\,32\mathrm{B})
\]

represents a pipeline using a 32B generator, a 0.6B critic, and a 32B refiner.

\subsection{Stage-wise Model Size Analysis Protocol}
\label{Stage-wise Model Size Analysis Protocol}

\begin{figure}[t]
    \centering
    \includegraphics[width=0.45\textwidth]{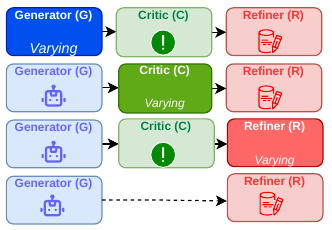}
    \caption{Stage-wise model size analysis. The three panels illustrate the generator, critic, and refiner sweeps. The highlighted stage is the stage under evaluation, whose model size is varied while the other two stages are held fixed. We also compare against a no-critique baseline.}
    \label{fig:scaling_matrix}
\end{figure}

Our objective is to quantify the contribution of model size at each stage of the self-refinement pipeline. To isolate the effect of an individual stage, we vary exactly one stage while keeping the remaining two stages fixed to the same models. This controlled design ensures that any performance differences can be attributed solely to the stage being varied (Figure \ref{fig:scaling_matrix}).

Specifically, we perform three stage-wise model size analysis experiments:

\begin{itemize}
    \item \textbf{Generator sweep:} vary the generator size while fixing the critic and refiner at the same size,
    \[
    P(G_i, C, R),
    \]
    where $G_i$ denotes the $i$-th generator model.

    \item \textbf{Critic sweep:} vary the critic size while fixing the generator and refiner at the same size,
    \[
    P(G, C_i, R),
    \]
    where $C_i$ denotes the $i$-th critic.

    \item \textbf{Refiner sweep:} vary the refiner size while fixing the generator and critic at the same size,
    \[
    P(G, C, R_i),
    \]
    where $R_i$ denotes the $i$-th refiner.
\end{itemize}

We repeat this procedure for every benchmark and every model family. For example, under the Qwen3-32B configuration, the generator sweep evaluates P(0.6B, 32B, 32B), P(1.7B, 32B, 32B), P(4B, 32B, 32B), P(8B, 32B, 32B), P(14B, 32B, 32B), and P(32B, 32B, 32B).

\subsection{No-Critique Baseline}
To determine whether explicit critique provides benefits beyond an additional refinement pass, we compare the standard self-refinement pipeline against a matched no-critique baseline. The no-critique pipeline is denoted by

\[
P(G,\emptyset,R),
\]

where the critic is removed and the refiner directly revises the generator output without receiving feedback.

To isolate the effect of introducing critique, we fix the critic to the smallest model in the Qwen family (Qwen3-0.6B) and compare it against the corresponding no-critique pipeline. For the generator sweep, the critique gain is defined as

\[
\Delta_{\mathrm{critique}}^{\mathrm{gen}}(G_i)
=
P(G_i,0.6\mathrm{B},R)
-
P(G_i,\emptyset,R),
\]

Similarly, for the refiner sweep,

\[
\Delta_{\mathrm{critique}}^{\mathrm{ref}}(R_i)
=
P(G,0.6\mathrm{B},R_i)
-
P(G,\emptyset,R_i),
\]

where only the refiner varies while the generator remains fixed.

For each benchmark, we report the average critique gain across the generator and refiner sweeps. By fixing the critic to Qwen3-0.6B, this comparison measures whether even a lightweight critic provides benefits beyond an additional refinement pass.

\section{Experimental Setup}

We conduct stage-wise model size analysis across five benchmarks spanning planning, summarization, logical reasoning, code optimization, and open-ended story-writing. Experiments are conducted with 2 model families (Qwen3 and Gemma 3) and 5 configurations (Qwen3-32B, Qwen3-14B, Qwen3-8B, Gemma 3-27B, and the no-critique baseline). For each benchmark, we independently vary the model assigned to one pipeline stage while holding the other two stages fixed, following the protocol described in Section \ref{Stage-wise Model Size Analysis Protocol}.

\begin{figure*}[t]
    \centering
    \includegraphics[width=1.0\textwidth]{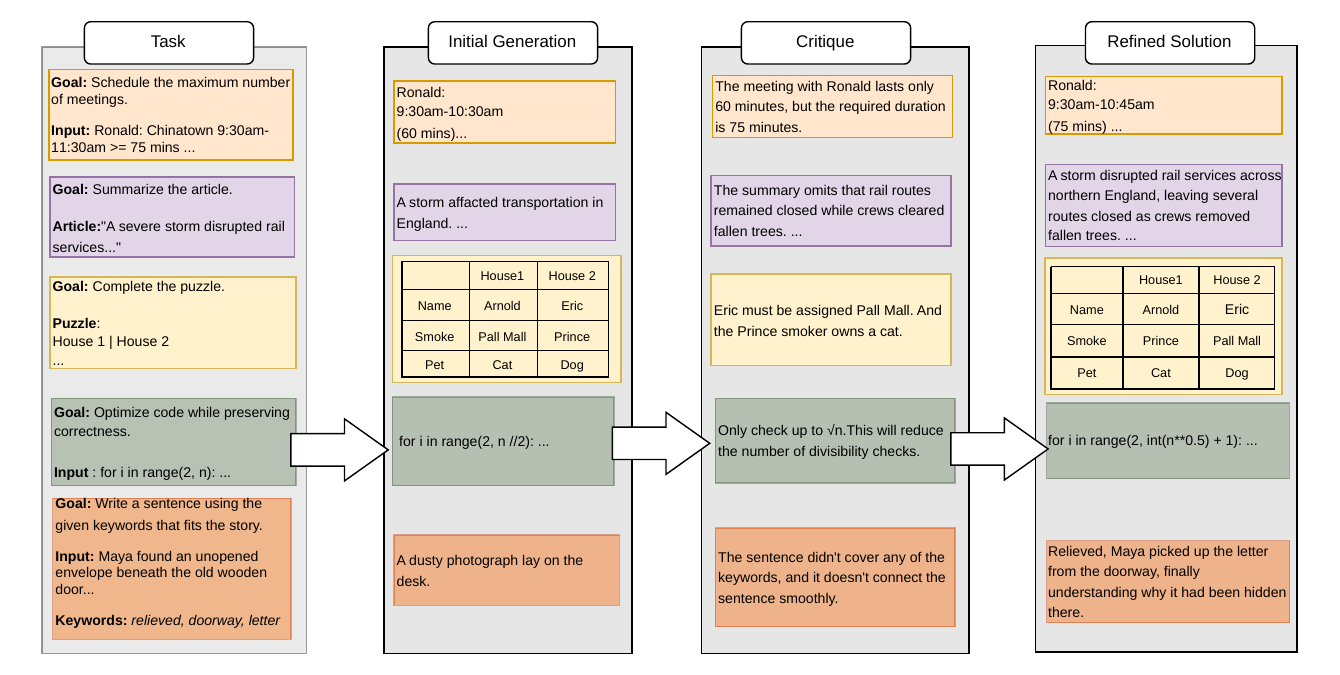}
    \caption{Representative examples from the five benchmarks. The first column briefly illustrates the objective and input for each task, followed by the initial generation, the generated critique, and the final refined solution, demonstrating how the self-refinement pipeline operates across diverse domains.}
    \label{fig:benchmark_examples}
\end{figure*}

\subsection{Models}

We evaluate our stage-wise scaling analysis using two families of open-weight large language models: Qwen3 and Gemma 3. These model families provide a diverse range of parameter scales while maintaining consistent architectures and training paradigms, making them suitable for studying the effect of model capacity across different stages of self-refinement.

For the Qwen3 family, we consider models with 0.6B, 1.7B, 4B, 8B, 14B, and 32B parameters. To examine whether our findings depend on the maximum available model capacity, we conduct three complete stage-wise scaling analyses using Qwen3-32B, Qwen3-14B, and Qwen3-8B as the largest models in the pipeline for the fixed stages. Unless otherwise specified, the main paper reports results from the Qwen3-32B configuration. The Qwen3-14B and Qwen3-8B experiments are presented in Appendix \ref{sec: Qwen14b_configuration} and Appendix \ref{sec: Qwen8_configuration} as robustness studies.

To further validate the generality of our observations, we repeat the analysis on the Gemma 3 family, using  Gemma 3-1B-IT, Gemma 3-4B-IT, Gemma 3-12B-IT, and Gemma 3-27B-IT. The corresponding results are reported in Appendix \ref{sec:gemma_trend_results} and exhibit the same qualitative stage-wise scaling trends observed for Qwen3.

\subsection{Benchmarks and Evaluation}
\label{eval_metric}
We evaluate stage-wise model size analysis on subsets of five diverse benchmarks spanning planning, summarization, logical reasoning, code optimization, and story generation. These tasks cover a broad range of reasoning and generation capabilities, allowing us to assess whether our findings generalize across heterogeneous application domains rather than being specific to a single task. Each benchmark is evaluated using its distinct task-specific metric. We put the subset details in Appendix \ref{dataset_specifics}.

\paragraph{Meeting Planning \citep{meetingplan}.}
In this task, the model must generate a feasible itinerary that satisfies a set of temporal and travel constraints. We evaluate planning performance using planning accuracy, defined as the percentage of generated itineraries that satisfy all constraints. An itinerary is considered correct only if it is fully feasible; any violation of the specified constraints results in a score of zero for that instance.

\paragraph{CNN/DailyMail \citep{hermann2015teaching, see2017get}.}
In this task, the model must generate a concise summary that captures the key information from the original news article. We evaluate performance using coverage rate, defined as the percentage of reference highlight sentences whose information is covered by the generated summary.

\paragraph{ZebraLogic \citep{Zebralogic}.}
In this task, the model must infer a complete logic grid puzzle that satisfies a set of logical constraints. We report cell accuracy, which measures the percentage of correctly assigned attribute values across all cells in the completed solution grid. 

\paragraph{PIE \citep{PIE}.}
In this task, the model must optimize a program to improve its efficiency while preserving its original functionality. We evaluate performance using \%OPT, defined as the percentage of generated programs that are both functionally correct and successfully optimized.

\paragraph{CollaboSentGen.} 
This is a benchmark derived from the ROCStories corpus \citep{commonsensestory}, where one of the middle three sentences is removed from a story and a set of keywords from the missing sentence is provided.
In this task, the model must generate a sentence using a given set of keywords that coherently fits into an existing story. We evaluate performance using Overall-Story-Fit, where GPT-4 \citep{gpt4} serves as the automatic evaluator to assess how well the generated sentence integrates with the surrounding narrative.

\paragraph{Stage-wise Sensitivity Metrics.}
To quantify the sensitivity of each pipeline stage to model size, we summarize each stage-wise sweep using two statistics: the performance range ($\Delta$), defined as the difference between the highest and lowest scores within a sweep, and the standard deviation ($\sigma$) of the performance scores. Larger values of $\Delta$ and $\sigma$ indicate that pipeline performance is more sensitive to model size increase.

\subsection{Prompting Strategy}
All three stages use task-specific few-shot instruction prompts for format and structure alignment \citep{languagemodelsfewshotlearners}. The complete prompts used for each benchmark are provided in Appendix \ref{prompts}.




Figure \ref{fig:benchmark_examples} illustrates the self-refinement pipeline on all five benchmarks before the quantitative analysis. These examples provide intuition for the different types of tasks considered in our evaluation before we analyze stage-wise scaling quantitatively.






\begin{figure*}[t]
    \centering
    \includegraphics[width=\textwidth]{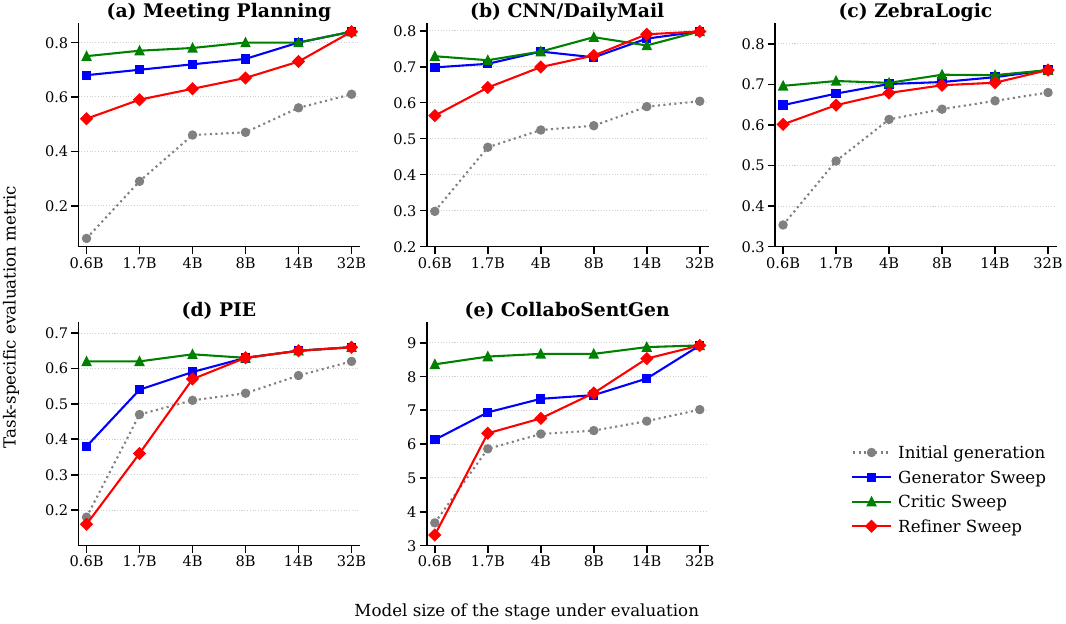}
    \caption{Stage-wise model-size sweeps across five benchmarks under the Qwen3-32B configuration. X-axis is the model sizes. Blue line, green line, and the red line represent Generator sweep, Critic sweep and the Refiner sweep. The grey dashed line reports the corresponding initial-generation performance without refinement. Steeper curves indicate greater sensitivity to the model size of that stage. Metrics are defined in Section \ref{eval_metric}.}
    \label{qwen32b_performance}
\end{figure*}

\begin{table}[t]
\centering
\scriptsize
\renewcommand{\arraystretch}{0.88}
\setlength{\tabcolsep}{3.5pt}
\setlength{\aboverulesep}{0.25ex}
\setlength{\belowrulesep}{0.25ex}

\begin{tabular}{llcccccc}
\toprule
\multirow{2}{*}{\textbf{Benchmark}} &
\multirow{2}{*}{\textbf{Model}} &
\multicolumn{2}{c}{\textbf{Generator}} &
\multicolumn{2}{c}{\textbf{Critic}} &
\multicolumn{2}{c}{\textbf{Refiner}} \\
\cmidrule(lr){3-4}
\cmidrule(lr){5-6}
\cmidrule(lr){7-8}
&
& $\Delta$ & Std. &
$\Delta$ & Std. &
$\Delta$ & Std. \\
\midrule
\multirow{2}{*}{Meeting Planning}
& Qwen  & 16.00 & 6.15 & 9.00 & 3.10 & 32.00 & 11.20 \\
& Gemma & 24.00 & 8.76 & 8.00 & 2.92 & 34.00 & 12.64 \\
\midrule
\multirow{2}{*}{CNN/DailyMail}
& Qwen  & 10.00 & 3.95 & 8.00 & 3.10 & 23.40 & 8.99 \\
& Gemma & 25.10 & 9.62 & 6.30 & 2.24 & 16.80 & 6.39 \\
\midrule
\multirow{2}{*}{ZebraLogic}
& Qwen  & 8.65 & 3.07 & 3.89 & 1.46 & 13.39 & 4.71 \\
& Gemma & 27.60 & 10.71 & 5.80 & 2.31 & 17.50 & 6.71 \\
\midrule
\multirow{2}{*}{PIE}
& Qwen  & 28.00 & 10.43 & 4.00 & 1.63 & 50.00 & 20.60 \\
& Gemma & 37.50 & 14.79 & 5.00 & 1.92 & 17.00 & 6.50 \\
\midrule
\multirow{2}{*}{CollaboSentGen}
& Qwen  & 2.79 & 0.96 & 0.56 & 0.20 & 5.61 & 2.03 \\
& Gemma & 2.46 & 1.038 & 0.63 & 0.26 & 4.36 & 1.92 \\
\bottomrule
\end{tabular}

\caption{Performance variability across model sizes when varying one stage while fixing the other two at the largest model. Values are reported in each benchmark's evaluation metric, as specified in Section~\ref{eval_metric}.}
\label{tab:stage_variability}
\end{table}

\begin{figure}[t]
    \centering
    \includegraphics[width=1.0\columnwidth, height=0.25\textheight]{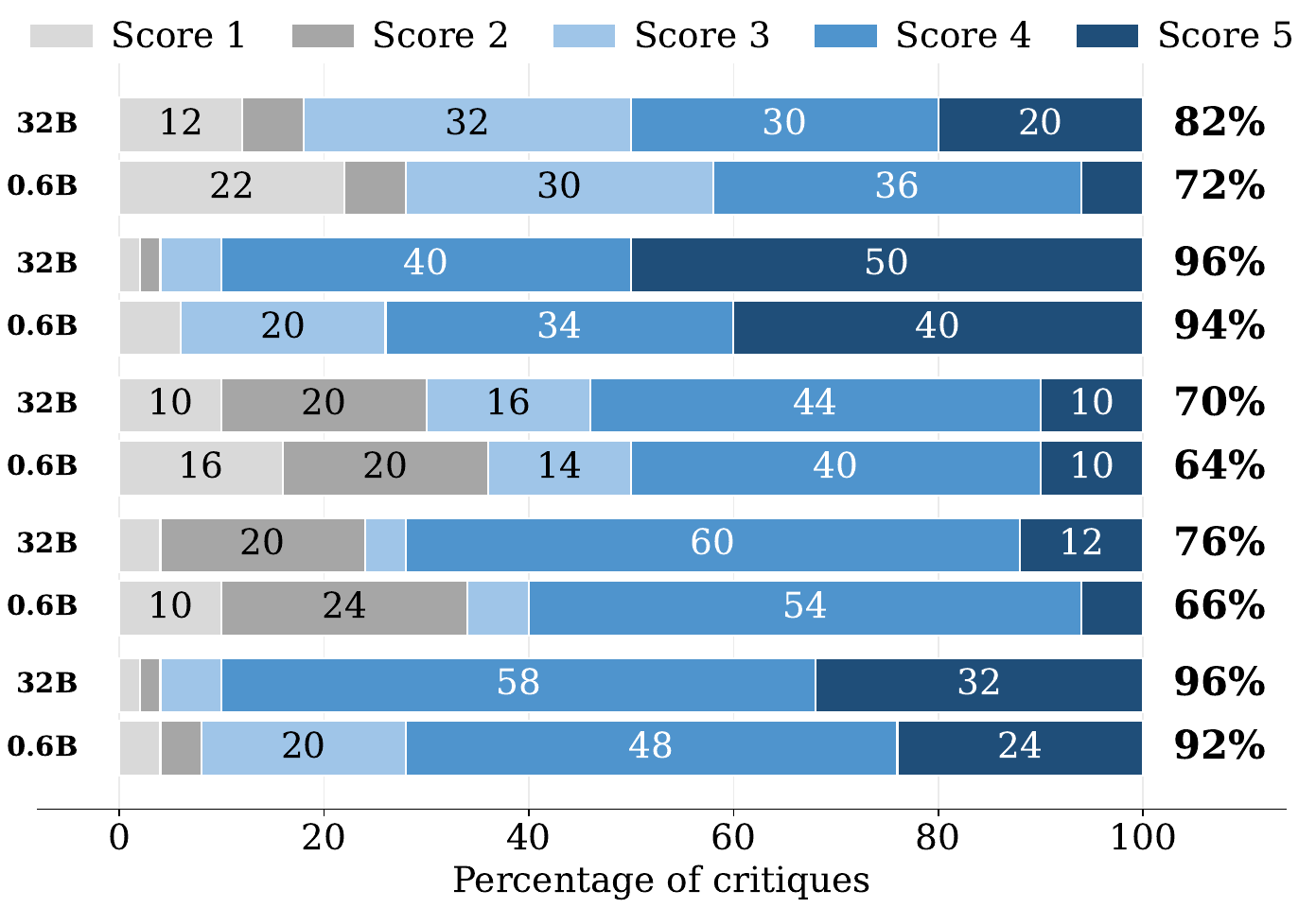}
    \caption{Distribution of manual critique-quality scores across five benchmarks. For each benchmark (top to bottom: Meeting Planning, CNN/DailyMail, ZebraLogic, PIE, and CollaboSentGen): Stacked segments show the percentage of critiques assigned scores 1–5, while the percentages on the right report the fraction of non-misleading critiques (scores 3–5).}
    \label{fig:critique_distribution}
\end{figure}

\section{Results}

\begin{table}[t]
\small
\centering
\renewcommand{\arraystretch}{1.1}

\begin{tabularx}{\columnwidth}{@{}p{0.28\columnwidth}X@{}}
\toprule
\textbf{Input} &
\textbf{Start:} The Castro (9:00 AM) \newline
Ronald: Chinatown (9:30--11:30, 75 min) \newline
Nancy: Nob Hill (11:15--12:00, 30 min) \\
\midrule
\textbf{Generated} &
09:30--10:30 Ronald \newline
10:50--11:20 Nancy \\
\midrule
\shortstack[l]{\textbf{0.6B Critic}\\\textbf{(Score 4)}} &
\textbf{Detects:} Ronald's meeting is only 60 min
(requires 75 min). \\
\midrule
\shortstack[l]{\textbf{32B Critic}\\\textbf{(Score 5)}} &
\textbf{Detects:} Ronald's meeting is only 60 min. \newline
\textbf{Detects:} Nancy is scheduled outside her
availability window. \\
\bottomrule
\end{tabularx}
\caption{Representative example illustrating performance of 0.6B critic and 32B critic.}
\label{tab:critic_example}
\end{table}

Figure ~\ref{qwen32b_performance} shows the effect of increasing model size at each stage independently while fixing the other two stages at Qwen3-32B. Each data point represents one pipeline. The blue line represents the trend of Generator sweep, the green line represents that of the Critic sweep, and the red line represents that of the Refiner sweep (defined in Section \ref{Stage-wise Model Size Analysis Protocol}). The gray dashed line represents generation without any refinement. Visually, a flatter line means the performance of the corresponding stage is less sensitive to the increased size of the model.

A consistent trend across all benchmarks is that scaling the size of the generator and the refiner substantially enhances the performance, which is visualized by the steeper lines in Figure \ref{qwen32b_performance} and quantitatively summarized in Table \ref{tab:stage_variability}. In contrast, the critic stage is relatively insensitive to model size, as larger critic models provide only marginal gains over smaller ones. Results for the Gemma 3-27B, Qwen3-14B, and Qwen3-8B configurations are provided in Appendix \ref{sec:gemma_trend_results}, \ref{sec: Qwen14b_configuration}, and \ref{sec: Qwen8_configuration}. They exhibit the same qualitative trends.

\subsection{Model Size Analysis of The Generator}
\paragraph{Finding 1. Pipeline performance improves consistently with generator size scaling.} Figure \ref{qwen32b_performance} shows that increasing the generator size consistently improves end-to-end pipeline performance across all five benchmarks. The same trend is reproduced in Figure \ref{gemma27b_performance}, \ref{qwen3_14b_performance}, and \ref{qwen3_8b_performance}, using different model configurations. 

We quantify this trend using the range ($\Delta$) and standard deviation (Std.) of performance across model sizes for each pipeline stage. The resulting statistics are summarized in Table~\ref{tab:stage_variability}. Under the Qwen3 family, the generator exhibits standard deviations ranging from 0.96 on CollaboSentGen to 10.43 percentage points on PIE, indicating that generator scaling substantially affects pipeline performance.

\subsection{Model Size Analysis of The Critic}
\paragraph{Finding 2.1 Pipeline performance is largely insensitive to critic size.}
This behavior is visually evident in Figure~\ref{qwen32b_performance}, where the critic curves are consistently flatter than those of the generator and refiner. We further quantify this trend by computing the range ($\Delta$) and standard deviation (Std.) of performance across model sizes for each stage (Table~\ref{tab:stage_variability}). Across all benchmarks and both model families, the critic consistently exhibits the lowest variability. For example, its standard deviation remains below 0.21 on CollaboSentGen and below 3.1 percentage points on all the other benchmarks, whereas generator and refiner show substantially larger variation. These results indicate that scaling the critic model size yields comparatively limited improvements in end-to-end performance.

To better understand this phenomenon, we investigate two possible explanations: (1) Critique quality changes only slightly across models, or (2) Even when larger critics produce higher-quality critiques, the refinement model may not fully utilize the additional information. We first examine the quality of the critiques themselves through a manual evaluation. Specifically, we randomly sample 50 critiques from the 0.6B and 32B critique models respectively from all five benchmarks. Following prior work on rubric-based evaluation \citep{rubric-evalution}, we design a task-specific five-point rubric to assess critique quality: A score of 5 denotes a completely correct, comprehensive, and well-justified critique; 4 indicates a correct critique with some omissions; 3 denotes an imperfect but non-misleading critique; 2 indicates partially misleading feedback containing at least one incorrect recommendation; and 1 denotes fundamentally incorrect or misleading feedback. We denote scores of 5 and 4 as correct, 3 as benign, and 2 and 1 as misleading.

Figure \ref{fig:critique_distribution} shows that the 0.6B and 32B critics rarely produce critiques with substantially different quality. When the 32B critic generates a correct critique, the 0.6B critic typically also produces a non-misleading critique (scores 3–5), although it may miss some errors. For example, on ZebraLogic, 70\% of the critiques produced by the 32B critic fall under scores 3-5 and 64\% of the critiques produced by 0.6B critic fall under scores 3-5.
Conversely, when the 32B critic produces misleading feedback (scores 1–2), the 0.6B critique is also usually misleading. The primary benefit of the larger critic is therefore detecting additional errors, rather than fundamentally changing whether the feedback is correct or incorrect. Table \ref{tab:critic_example} is a representative example. 32B model detects more errors than 0.6B model but they are both providing correct feedback. Although the 32B critic often provides more complete feedback, the corresponding improvement in refinement performance remains modest (Figure \ref{qwen32b_performance}). This suggests that the refiner does not always translate the additional critique information into better revisions, providing evidence for the second explanation as well.

\begin{table}[t]
\centering
\small
\begin{tabular*}{\columnwidth}{@{\extracolsep{\fill}}lcc}
\toprule
\textbf{Benchmark} & \textbf{Generator Sweep} & \textbf{Refiner Sweep} \\
\midrule
Meeting Planning & +9.67 & +10.34 \\
CNN/DailyMail    & +4.56  & +5.21  \\
ZebraLogic       & +7.97 & +7.86  \\
PIE              & +9.03 & +11.06 \\
CollaboSentGen   & +3.71     & +4.01     \\
\bottomrule
\end{tabular*}
\caption{Average critique gain with Qwen3-0.6B as the critic over the matched no-critique baseline across the generator and refiner sweeps. }
\label{tab:critique_gain}
\end{table}





\paragraph{Finding 2.2 Even the smallest critic outperforms no-critique pipeline}
To determine whether the improvements of self-refinement arise from the critique itself or simply from performing an additional inference step, we compare the standard pipeline P(G, 0.6B, R) with the smallest critic to a matched no-critique baseline P(G, $\varnothing$, R), in which the refiner directly revises the generator output without receiving critique.


We observe that even incorporating the smallest explicit critic would yield improvements over the initial generation across all five benchmarks, as shown in Table \ref{tab:critique_gain}. On average, using Qwen-0.6b critic improves performance by 9.67 and 10.34 percentage points on Meeting Planning benchmark for the generator and refiner sweeps. We observe similar gains on the remaining benchmarks. These results demonstrate that the benefit of self-refinement cannot be attributed solely to additional computation. Instead, explicit critique provides complementary guidance that enables the refiner to produce higher-quality outputs. 
\begin{figure}[!t]
    \centering
    \includegraphics[width=0.8\columnwidth]{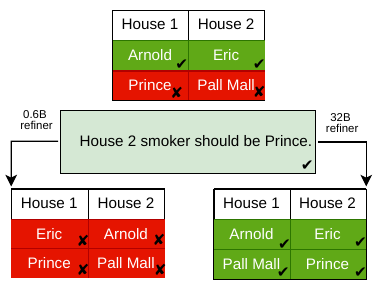}
    \caption{A degradation example from ZebraLogic. Despite a correct critique, the 0.6B refiner is introducing new errors while the 32B refiner is incorporating the critique.}
    \label{degradation_example}
\end{figure}

\subsection{Model Size Analysis of The Refiner}
\paragraph{Finding 3.1 Pipeline performance is highly sensitive to refiner size.}

The steeper lines of the refiner in Figure \ref{qwen32b_performance} show that across all five benchmarks, the refiner consistently exhibits substantially greater variability than the generator and critic. This observation is quantified in Table \ref{tab:stage_variability}.  For example, under the Qwen3 family, the refiner reaches a standard deviation of 20.60 percentage points on PIE compared to only 1.63 percentage points for the critic, 11.20 percentage points versus 3.10 percentage points on Meeting Planning, and 8.99 percentage points versus 3.10 percentage points on CNN/DailyMail.

\textbf{Finding 3.2. Weak refiners can degrade initial performance.}
\label{weak_refiner_hurting_performance}
Under the Qwen3-32B configuration, 12 of the 30 evaluated refiner pipelines (5 benchmarks × 6 refiner sizes) perform worse than the corresponding initial generation. The complete pipelines of all degraded configurations are in Appendix \ref{degraded_pipeline_table}. To understand the underlying failure mode, we manually analyzed 50 degradation events from the most extreme case, P(32B,32B,0.6B), where the weakest refiner exhibits the greatest performance drop. In all 50 events, the 0.6B refiner produced a worse output than both the initial generation and the 32B refiner under the same initial solution and critique.

Among the 50 degradation events, 41 occur despite receiving non-misleading critiques (human scores 3--5). In these cases, the 0.6B refiner unnecessarily modifies correct portions of the original solution while attempting to apply the critique, whereas the 32B refiner generally preserves correct content and performs only the required edits. The remaining 9 cases involve misleading critiques (scores 1--2). The 0.6B refiner follows the incorrect feedback, propagating critique errors into the final solution, while the 32B refiner ignores the misleading suggestions and preserves the original solution. Figure \ref{degradation_example} presents a representative degradation example. Both refiners receive the same critique, yet they respond differently. The 0.6B refiner unnecessarily modifies previously correct assignments, introducing new errors beyond those identified in the critique. In contrast, the 32B refiner performs only the required edits and preserves the remaining solution. This example reflects the dominant failure mode observed during manual analysis.

\section{Conclusion}
We presented the first stage-wise model size analysis of self-refinement pipelines. Across 5 benchmarks, 2 model families, and 4 configurations, we showed that pipeline performance is highly sensitive to the capacities of the generator and refiner, but comparatively insensitive to the critic size. While larger critics provide only modest gains, even a lightweight critic consistently outperforms a no-critique pipeline. In the meantime, a weak refiner can hurt the initial performance. These findings suggest that model capacity should be allocated non-uniformly across self-refinement pipelines, providing practical guidance for designing more compute-efficient multi-stage language model systems and LLM agentic systems.


\section*{Limitations}

This work analyzes stage-wise model scaling in the canonical generate–critique–refine pipeline with a single refinement iteration. Our findings may not directly generalize to more complex agentic systems that incorporate additional components such as retrieval, planning, tool use, memory, or multiple refinement rounds. Furthermore, although we evaluate two model families (Qwen3 and Gemma 3) across five diverse benchmarks, extending the analysis to additional architectures, tasks, and modalities would further establish the generality of the observed scaling trends. Finally, our study focuses on model capacity allocation and does not investigate other forms of inference-time scaling, such as adaptive routing, increased decoding budgets, or iterative refinement beyond a single critique–refine cycle.


\bibliography{custom}
\clearpage
\appendix

\appendix
\section{Implementation details}
\label{sec: implementation_details}
All experiments were conducted using Qwen3 and Gemma 3 models on NVIDIA H100 GPUs. We used task-specific few-shot prompts for the generator, critic, and refiner stages, with the complete prompt templates provided in the appendix. Generation was performed using greedy decoding with a temperature of 0.0 and a maximum output length of 700 new tokens.  The same decoding configuration was used across model sizes and pipeline stages to ensure that observed performance differences could be attributed to model capacity rather than changes in inference settings.

\section{Dataset Specifics}
\label{dataset_specifics}
We evaluate on a subset of each benchmark to keep the experiments computationally tractable. Depending on the benchmark, we either randomly sample evaluation instances or restrict the evaluation to subsets of controlled difficulty, as described below.

\paragraph{Meeting Planning.}
We evaluate on 500 examples. Since even the largest model used in our experiments struggles on the most difficult instances \citep{meetingplan}, we restrict the evaluation to problems involving at most five people.

\paragraph{CNN/DailyMail \citep{hermann2015teaching, see2017get}.}
We evaluate on the first 500 examples from version 1.0.0 of the official test split.

\paragraph{ZebraLogic.}
We evaluate on 320 examples. Since even the largest model used in our experiments struggles on the most difficult puzzles \citep{Zebralogic}, we restrict the evaluation to puzzles with at most 10 cells.

\paragraph{PIE \citep{PIE}.}
Because evaluating each generated program is computationally expensive, we randomly sample 500 examples from the benchmark.

\paragraph{CollaboSentGen \citep{commonsensestory}.}
We evaluate on 500 randomly selected examples.

\section{Additional Results on the Gemma Family}
\label{sec:gemma_trend_results}
\begin{figure*}[!t]
    \centering
    \includegraphics[width=\textwidth]{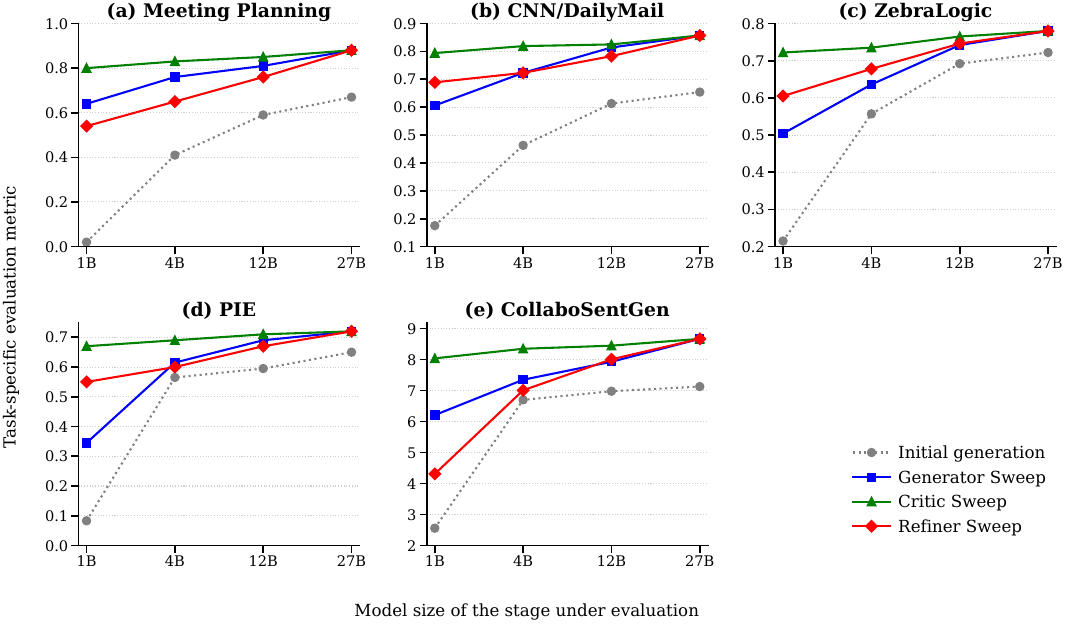}
    \caption{Stage-wise model-size sweeps across five benchmarks under the Gemma 3-27 configuration. X-axis is the model sizes. Blue line, green line, and the red line represent Generator sweep, Critic sweep and the Refiner sweep respectively. The grey dashed line reports the corresponding initial-generation performance without refinement. Steeper curves indicate greater sensitivity to the model size of that stage. Metrics are defined in Section \ref{eval_metric}}
    \label{gemma27b_performance}
\end{figure*}

To evaluate whether our observations generalize beyond Qwen3, we repeat the stage-wise scaling analysis using the Gemma family. Specifically, we evaluate Gemma 3 1B-IT, Gemma 3 4B-IT, Gemma 3 12B-IT, and Gemma 3 27B-IT. We plotted the results in Figure \ref{gemma27b_performance}. The Gemma results closely mirror those observed for Qwen3. Across all five benchmarks, pipeline performance is largely sensitive to both the size of the generator and the refiner. A weak refiner can underperform the initial generation. The performance is the least sensitive to critic size. These results demonstrate that our conclusions are not specific to a particular model family.

\section{Qwen3-14B Configuration}
\label{sec: Qwen14b_configuration}
\begin{figure*}[!t]
    \centering
    \includegraphics[width=\textwidth]{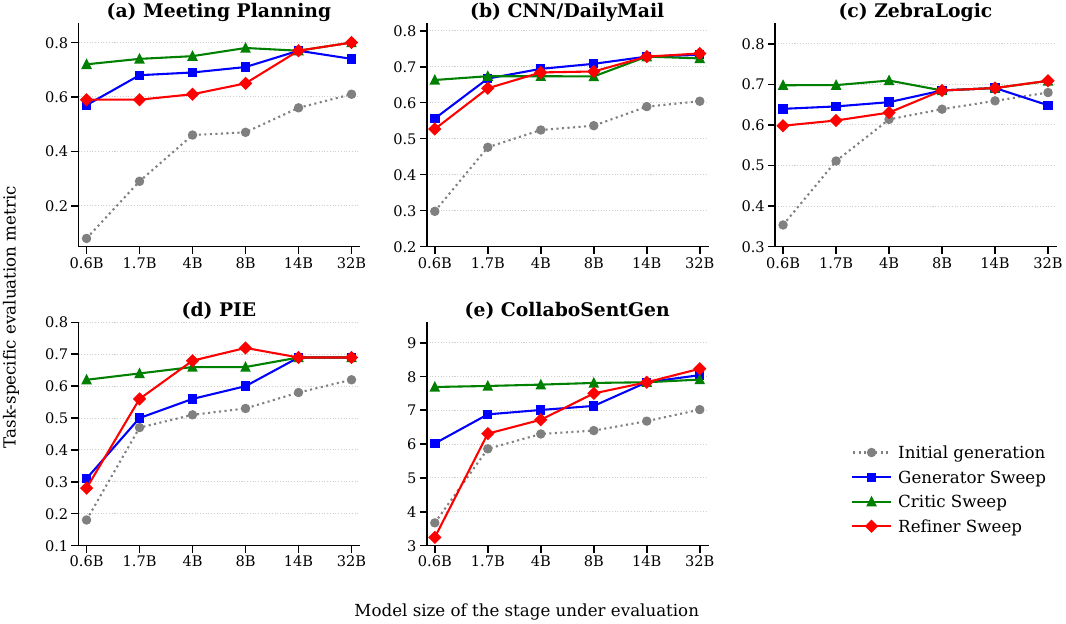}
        \caption{Stage-wise model-size sweeps across five benchmarks under the Qwen3-14B configuration. X-axis is the model sizes. Blue line, green line, and the red line represent Generator sweep, Critic sweep and the Refiner sweep respectively. The grey dashed line reports the corresponding initial-generation performance without refinement. Steeper curves indicate greater sensitivity to the model size of that stage. Metrics are defined in Section \ref{eval_metric}}
    \label{qwen3_14b_performance}
\end{figure*}

Figure~\ref{qwen3_14b_performance} compares the effect of increasing model size at each stage independently while fixing the other two stages at Qwen3-14B. The blue line represents the trend of Generator sweep, the green line represents that of the Critic sweep, and the red line represents that of the Refiner sweep (Defined in Section \ref{Stage-wise Model Size Analysis Protocol}) The same qualitative observations reported in the main text remain consistent. In particular, generator and refiner scaling substantially affect pipeline performance, whereas critique scaling produces comparatively smaller gains.

\section{Qwen3-8B Configuration}
\label{sec: Qwen8_configuration}
\begin{figure*}[!t]
    \centering
    \includegraphics[width=\textwidth]{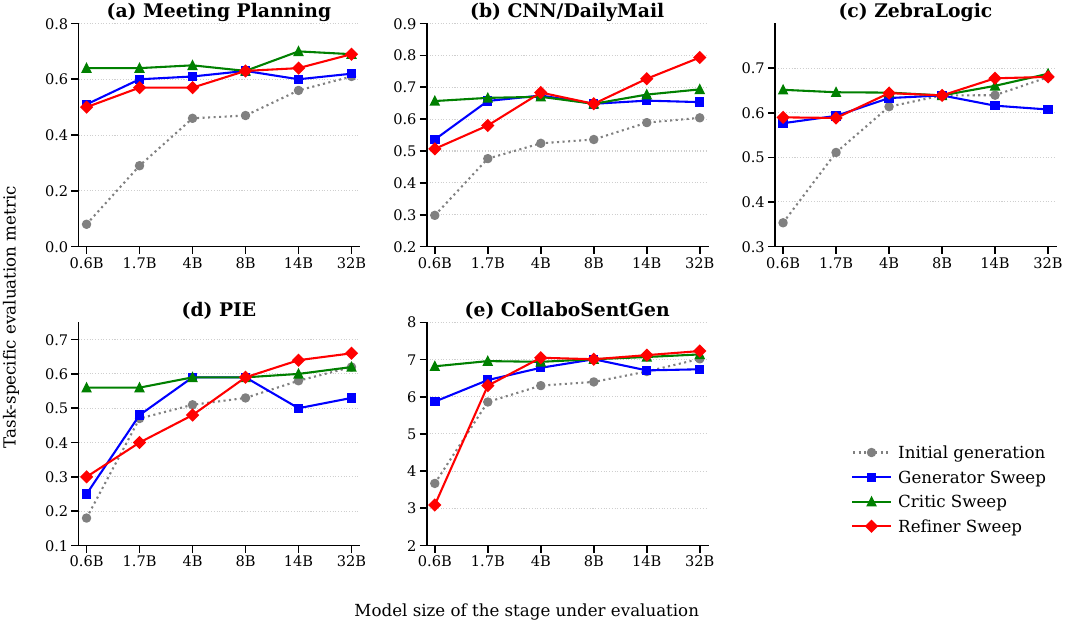}
       \caption{Stage-wise model-size sweeps across five benchmarks under the Qwen3-8B configuration. X-axis is the model sizes. Blue line, green line, and the red line represent Generator sweep, Critic sweep and the Refiner sweep respectively. The grey dashed line reports the corresponding initial-generation performance without refinement. Steeper curves indicate greater sensitivity to the model size of that stage. Metrics are defined in Section \ref{eval_metric}}
    \label{qwen3_8b_performance}
\end{figure*}

Figure~\ref{qwen3_8b_performance} compares the effect of increasing model size at each stage independently while fixing the other two stages at Qwen3-8B. The blue line represents the trend of Generator sweep, the green line represents that of the Critic sweep, and the red line represents that of the Refiner sweep (Defined in Section \ref{Stage-wise Model Size Analysis Protocol}) The same qualitative observations reported in the main text remain consistent. In particular, generator and refiner scaling substantially affect pipeline performance, whereas critique scaling produces comparatively smaller gains.

\section{Degradation Pipelines}
\label{degraded_pipeline_table}
To complement the qualitative analysis presented in Section \ref{weak_refiner_hurting_performance}, Table \ref{tab:degradation_summary} summarizes all benchmark--refiner-size combinations in which the final refined output performs worse than the corresponding initial generation. These degradation cases occur across multiple benchmarks and model families, indicating that the phenomenon is systematic rather than limited to a single task or model. This table reports the refiner sizes that lead to degradation under each experimental configuration.

\clearpage

\begin{table*}[t]
\centering

\begin{subtable}[t]{0.48\textwidth}
\centering
\caption{Qwen3-32B}
\small
\begin{tabular}{lcccccc}
\toprule
Benchmark & 0.6B & 1.7B & 4B & 8B & 14B & 32B \\
\midrule
Meeting Planning & \checkmark & \checkmark &  &  &  &  \\
CNN/DailyMail    & \checkmark &            &  &  &  &  \\
ZebraLogic       & \checkmark & \checkmark & \checkmark &  &  &  \\
PIE              & \checkmark & \checkmark & \checkmark & \checkmark &  &  \\
CollaboSentGen   & \checkmark & \checkmark &  &  &  &  \\
\midrule
Total & \multicolumn{6}{c}{12 / 30} \\
\bottomrule
\end{tabular}
\end{subtable}
\hfill
\begin{subtable}[t]{0.48\textwidth}
\centering
\caption{Qwen3-14B}
\small
\begin{tabular}{lcccccc}
\toprule
Benchmark & 0.6B & 1.7B & 4B & 8B & 14B & 32B \\
\midrule
Meeting Planning & \checkmark &  &  &  &  &  \\
CNN/DailyMail    & \checkmark &  &  &  &  &  \\
ZebraLogic       & \checkmark & \checkmark & \checkmark &  &  &  \\
PIE              & \checkmark & \checkmark &  &  &  &  \\
CollaboSentGen   & \checkmark & \checkmark &  &  &  &  \\
\midrule
Total & \multicolumn{6}{c}{9 / 30} \\
\bottomrule
\end{tabular}
\end{subtable}

\vspace{1em}

\begin{subtable}[t]{0.48\textwidth}
\centering
\caption{Qwen3-8B}
\small
\begin{tabular}{lcccccc}
\toprule
Benchmark & 0.6B & 1.7B & 4B & 8B & 14B & 32B \\
\midrule
Meeting Planning &  &  &  &  &  &  \\
CNN/DailyMail    & \checkmark &  &  &  &  &  \\
ZebraLogic       & \checkmark & \checkmark &  &  &  &  \\
PIE              & \checkmark & \checkmark & \checkmark &  &  &  \\
CollaboSentGen   & \checkmark & \checkmark &  &  &  &  \\
\midrule
Total & \multicolumn{6}{c}{8 / 30} \\
\bottomrule
\end{tabular}
\end{subtable}
\hfill
\begin{subtable}[t]{0.48\textwidth}
\centering
\caption{Gemma3-27B}
\small
\begin{tabular}{lcccc}
\toprule
Benchmark & 1B & 4B & 12B & 27B \\
\midrule
Meeting Planning & \checkmark & \checkmark &  &  \\
CNN/DailyMail    &            &            &  &  \\
ZebraLogic       & \checkmark & \checkmark &  &  \\
PIE              & \checkmark & \checkmark &  &  \\
CollaboSentGen   & \checkmark & \checkmark &  &  \\
\midrule
Total & \multicolumn{4}{c}{8 / 20} \\
\bottomrule
\end{tabular}
\end{subtable}
\caption{Refiner sizes that perform below the corresponding initial generation under different model configurations. A checkmark indicates that using the corresponding refiner size results in lower performance than the initial generation.}
\label{tab:degradation_summary}
\end{table*}

\section{Prompts}
\label{prompts}
\paragraph{Generator Prompts}The generator receives the task description with examples ($x$) and produces an initial solution $y_0$. 

\paragraph{Critic Prompts}The critic receives the task description with examples ($x$) and the initial solution $y_0$. It analyzes the generated solution $y_0$, identify errors, and provide actionable critique $f$. 

\paragraph{Refiner Prompts} The refiner receives the task description with examples ($x$), the initial solution $y_0$, and the critique $f$, and is instructed to produce a revised final solution $y_r$ that incorporates the critique. 

\begin{figure*}[!t]
    \centering
    \includegraphics[width=\textwidth]{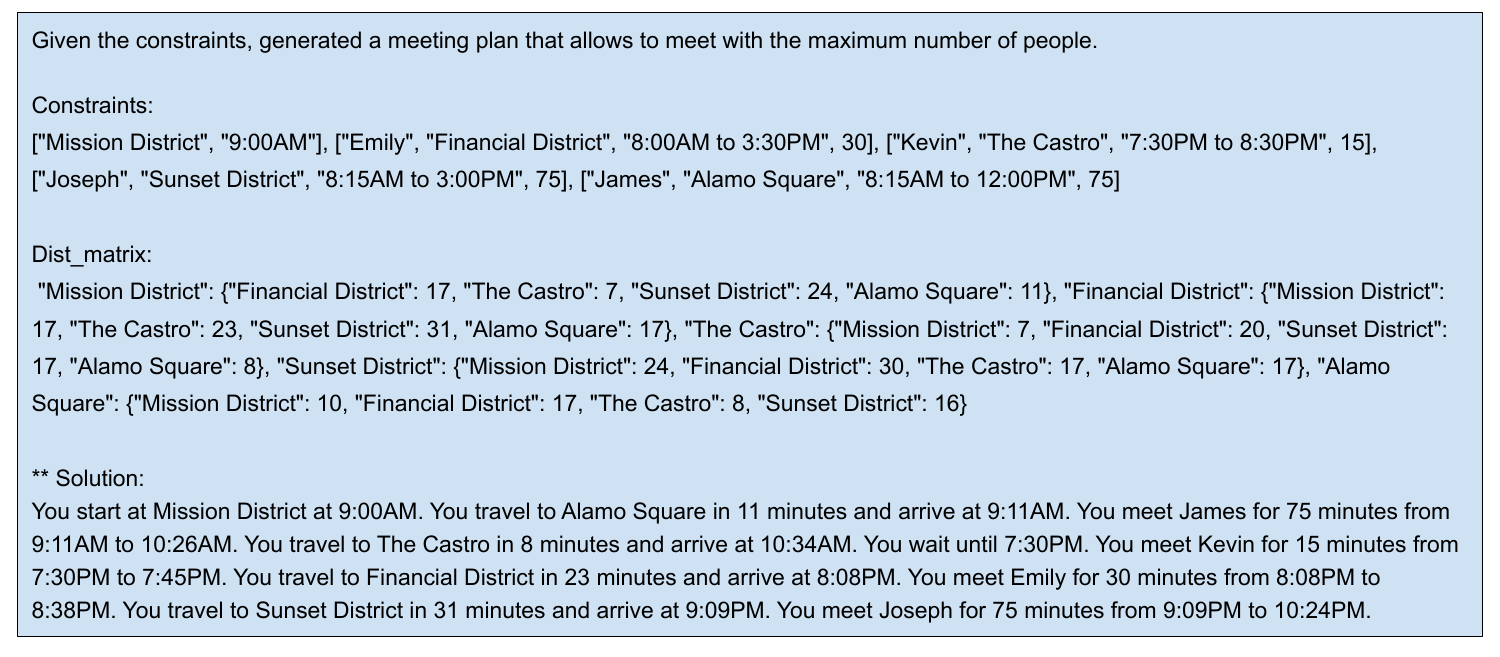}
    \caption{Generator prompt for Meeting Plan}
    \label{prompts_meeting_plan_gen}
\end{figure*}

\begin{figure*}[!t]
    \centering
    \includegraphics[width=\textwidth]{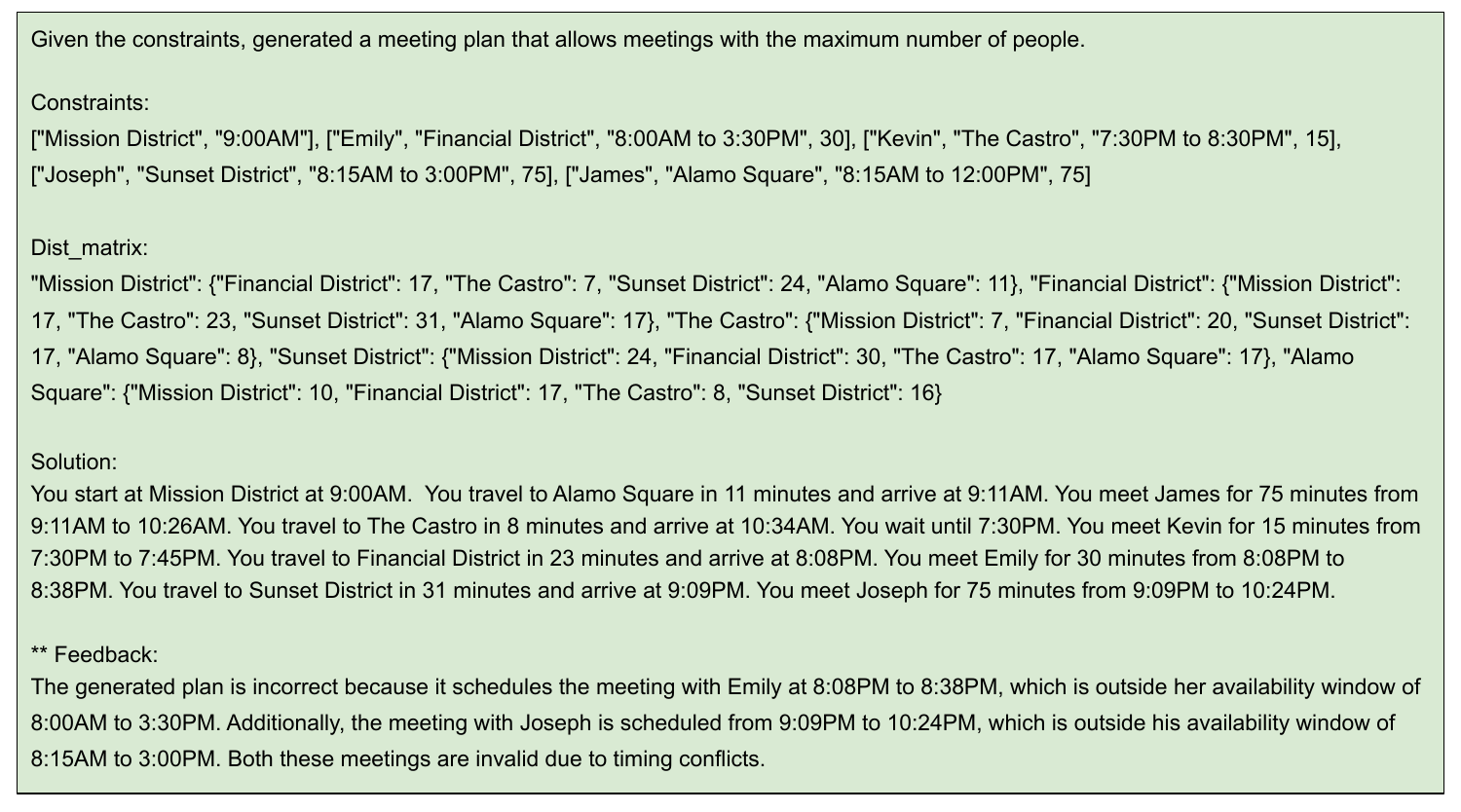}
    \caption{Critic prompt for Meeting Plan}
    \label{prompts_meeting_plan_critic}
\end{figure*}

\begin{figure*}[!t]
    \centering
    \includegraphics[width=\textwidth]{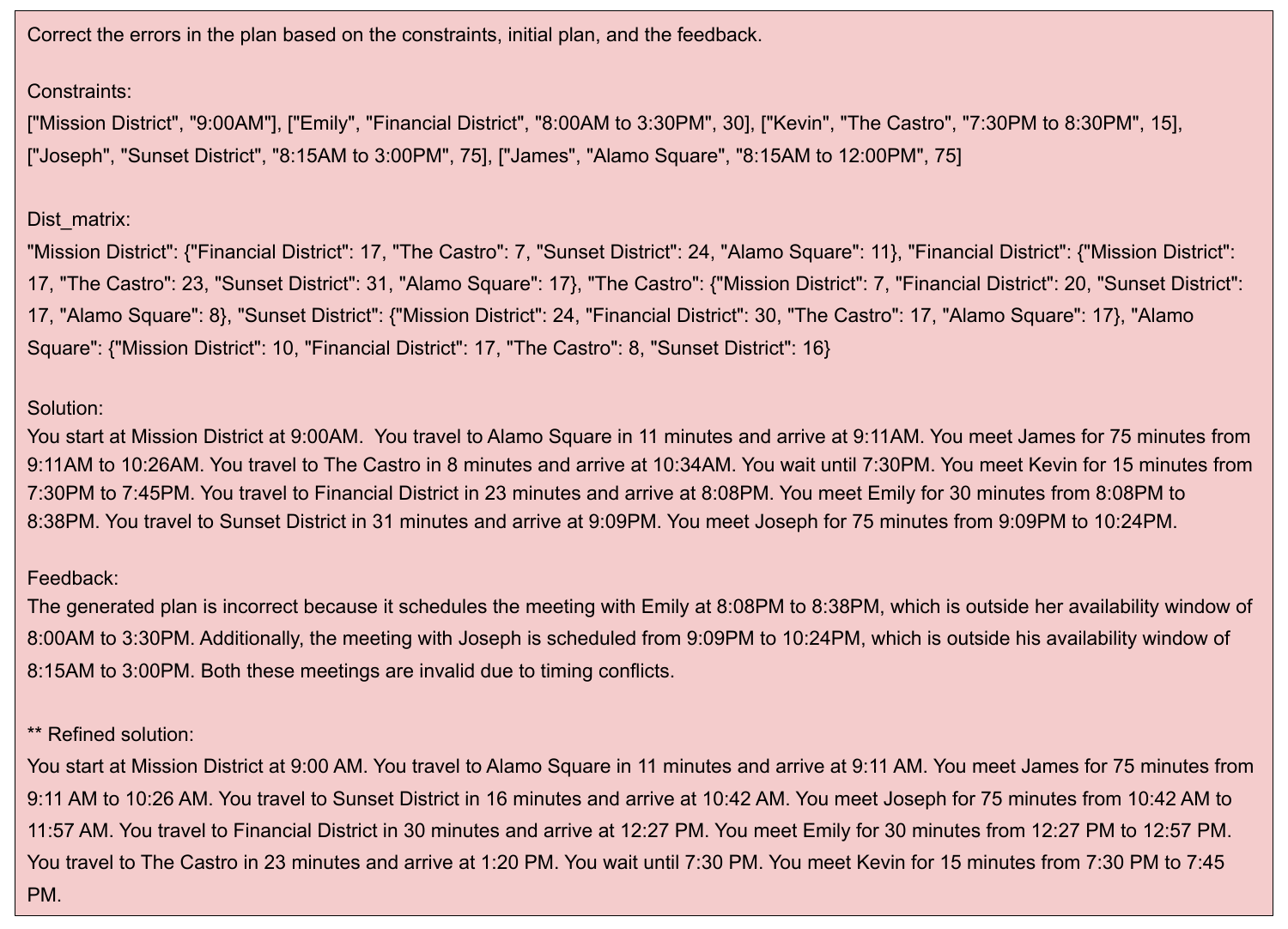}
    \caption{Refiner prompt for Meeting Plan}
    \label{prompts_meeting_plan_refiner}
\end{figure*}

\begin{figure*}[!t]
    \centering
    \includegraphics[width=\textwidth]{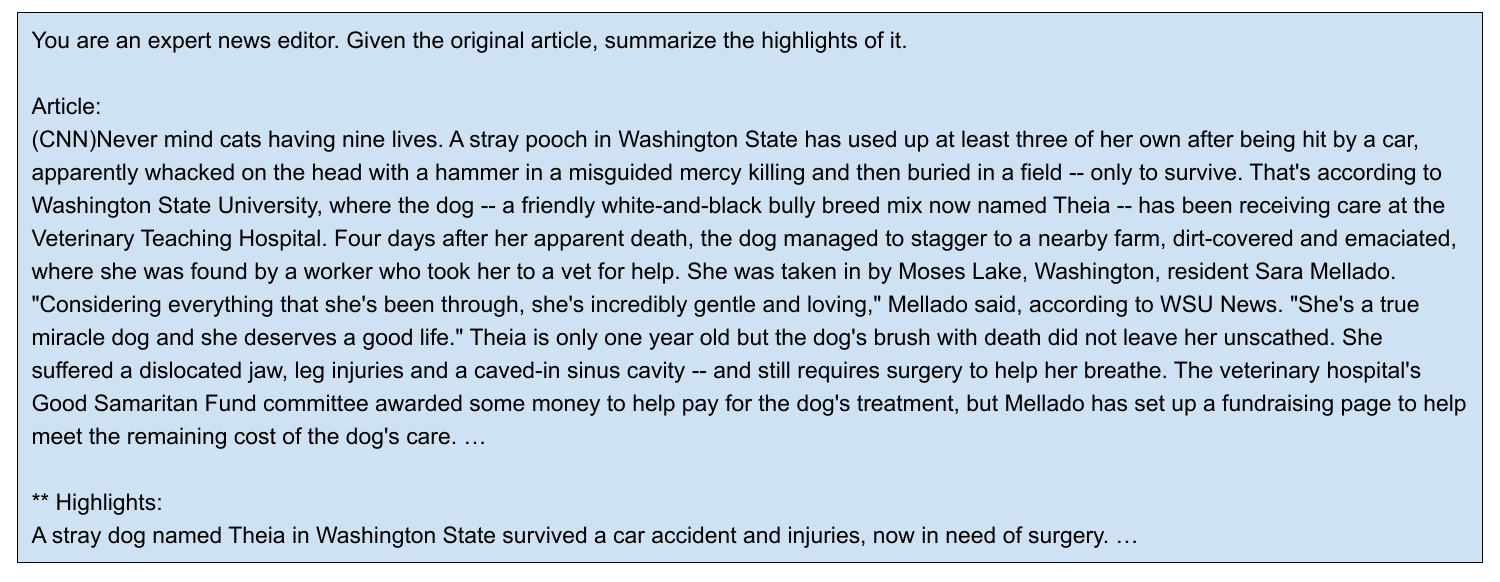}
    \caption{Generator prompt for cnn/DailyMail}
    \label{prompts_cnnDaily_gen}
\end{figure*}

\begin{figure*}[!t]
    \centering
    \includegraphics[width=\textwidth]{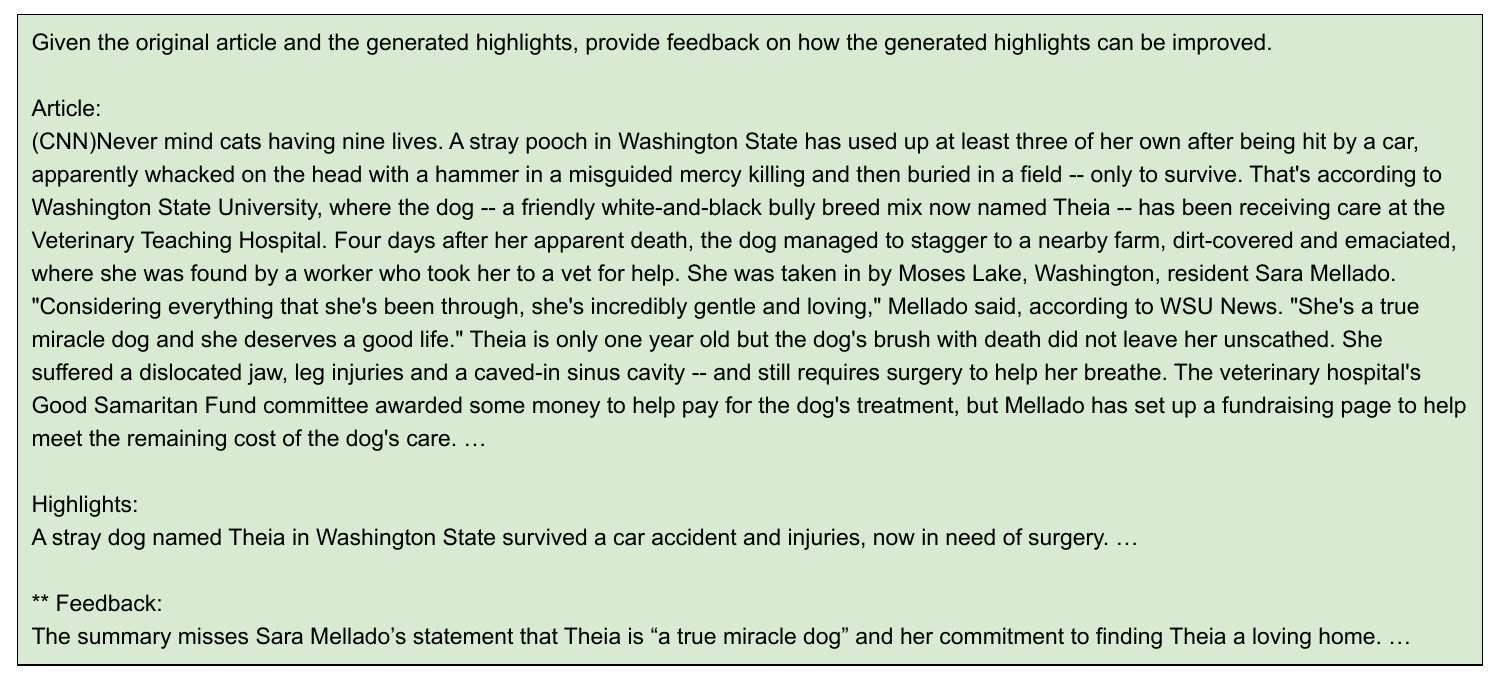}
    \caption{Critic prompt for cnn/DailyMail}
    \label{prompts_cnnDaily_critic}
\end{figure*}

\begin{figure*}[!t]
    \centering
    \includegraphics[width=\textwidth]{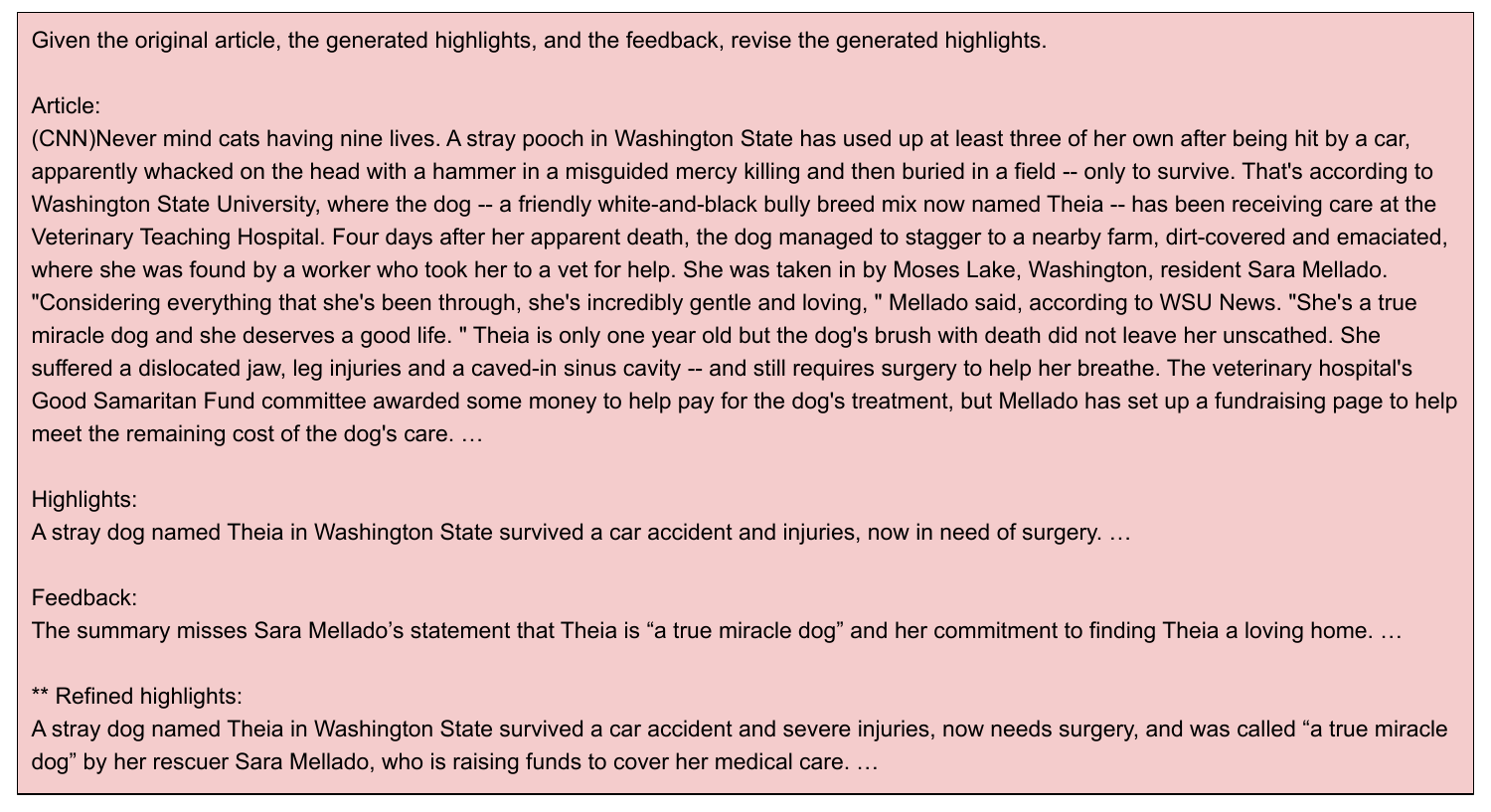}
    \caption{Refiner prompt for cnn/DailyMail}
    \label{prompts_cnnDaily_refiner}
\end{figure*}

\begin{figure*}[!t]
    \centering
    \includegraphics[width=\textwidth]{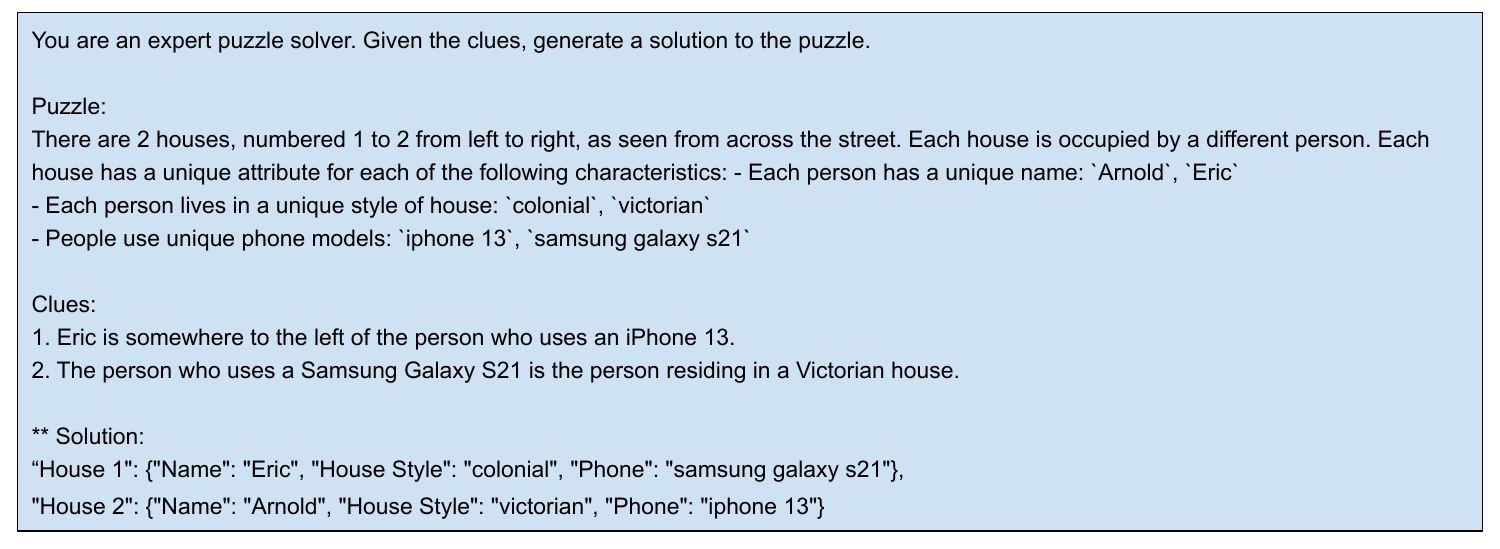}
    \caption{Generator prompt for ZebraLogic}
    \label{prompts_zebralogic_gen}
\end{figure*}

\begin{figure*}[!t]
    \centering
    \includegraphics[width=\textwidth]{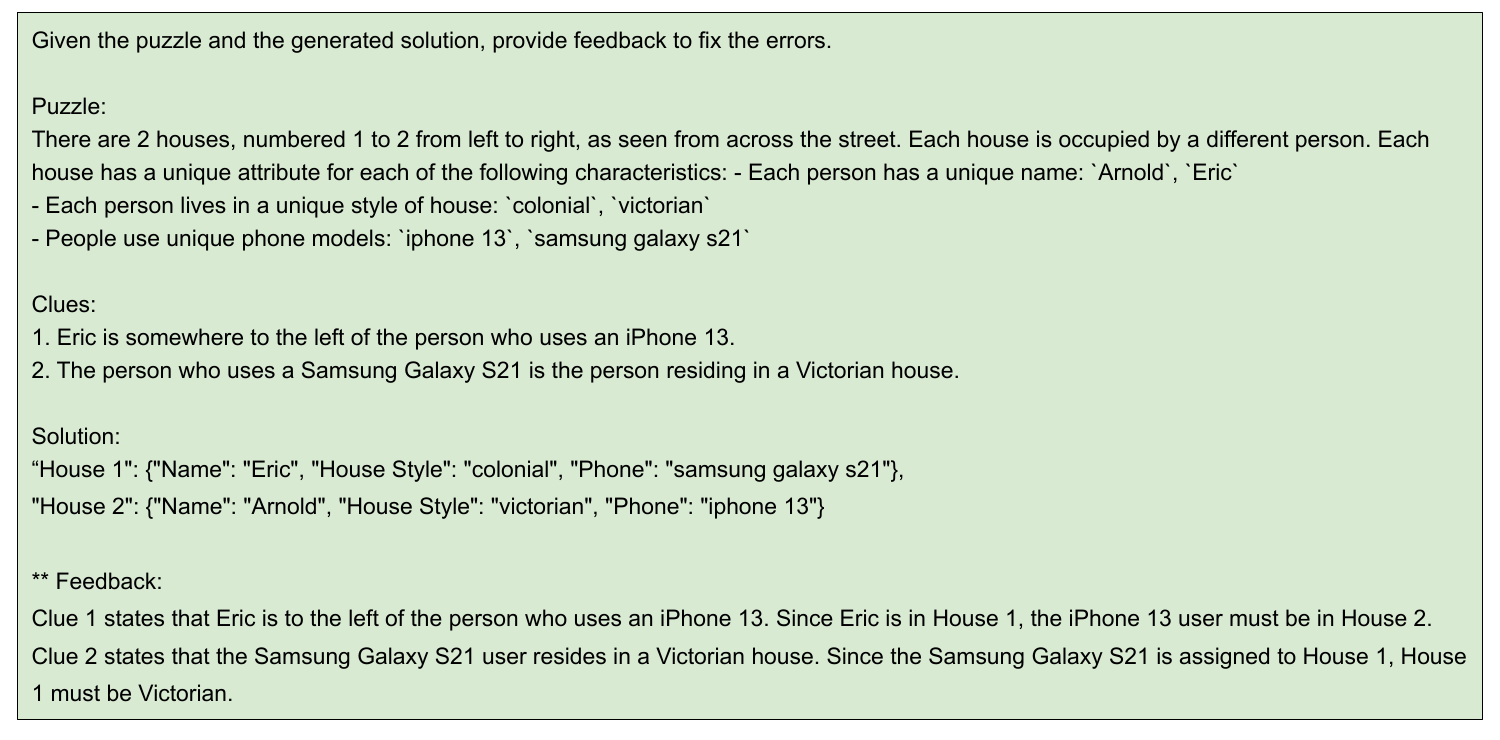}
    \caption{Critic prompt for ZebraLogic}
    \label{prompts_ZebraLogic_critic}
\end{figure*}

\begin{figure*}[!t]
    \centering
    \includegraphics[width=\textwidth]{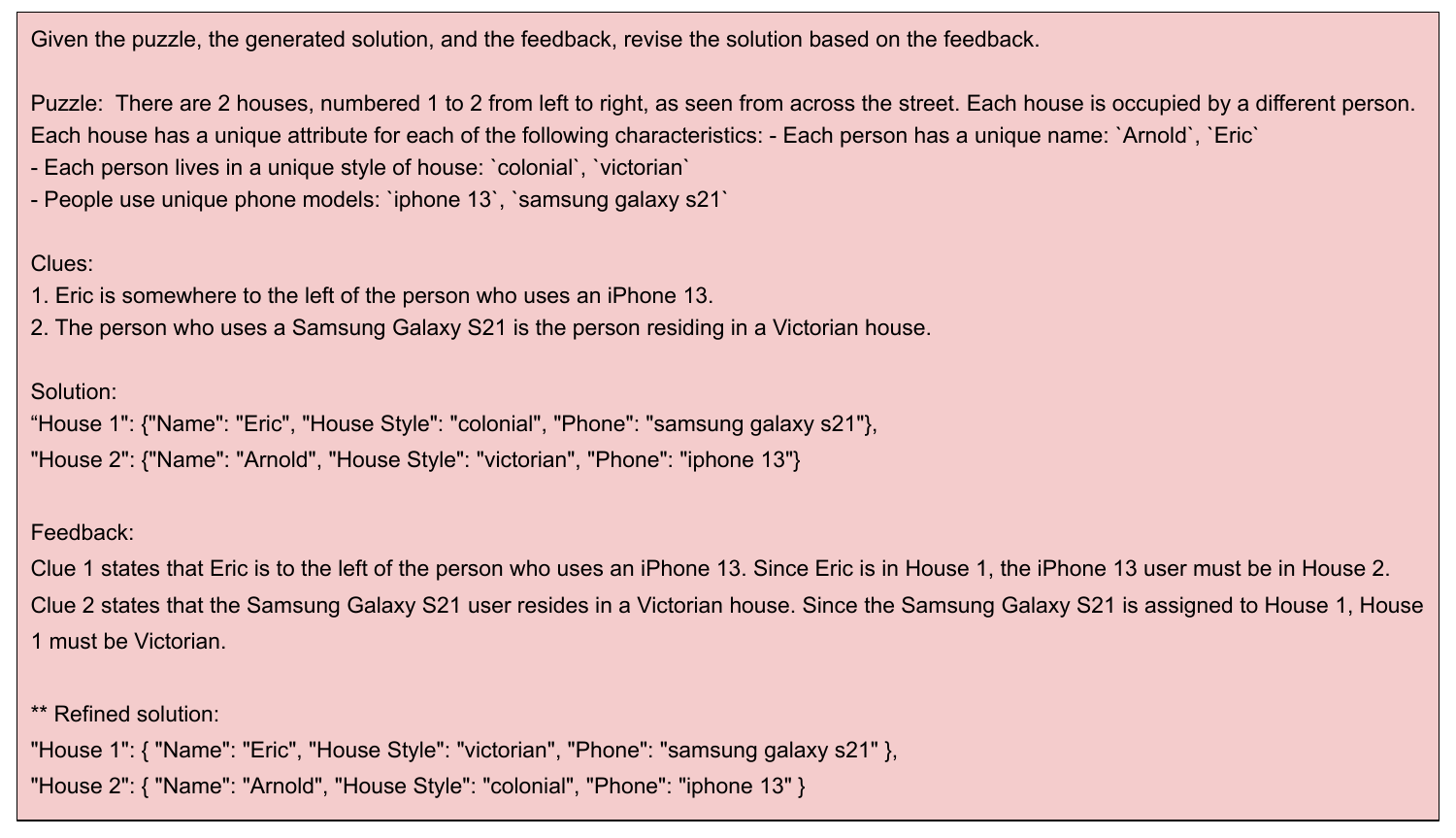}
    \caption{Refiner prompt for ZebraLogic}
    \label{prompts_zebralogic_refiner}
\end{figure*}

\begin{figure*}[!t]
    \centering
    \includegraphics[width=\textwidth]{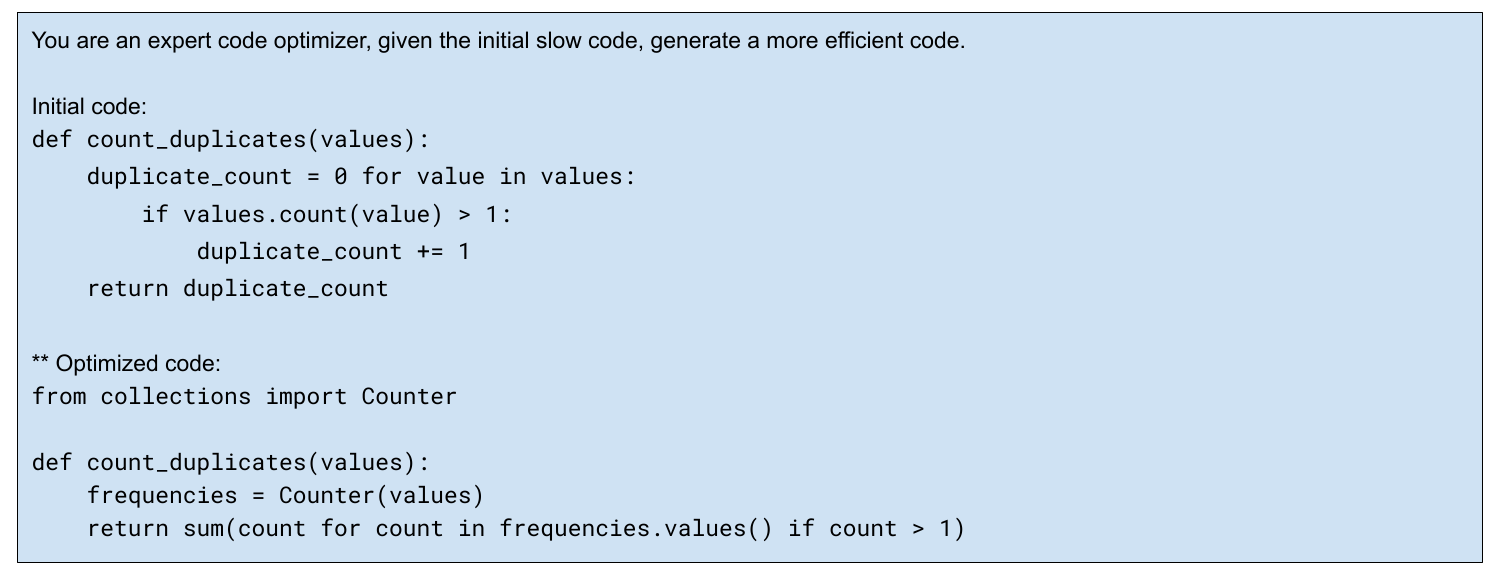}
    \caption{Generator prompt for PIE}
    \label{prompts_PIE_gen}
\end{figure*}

\begin{figure*}[!t]
    \centering
    \includegraphics[width=\textwidth]{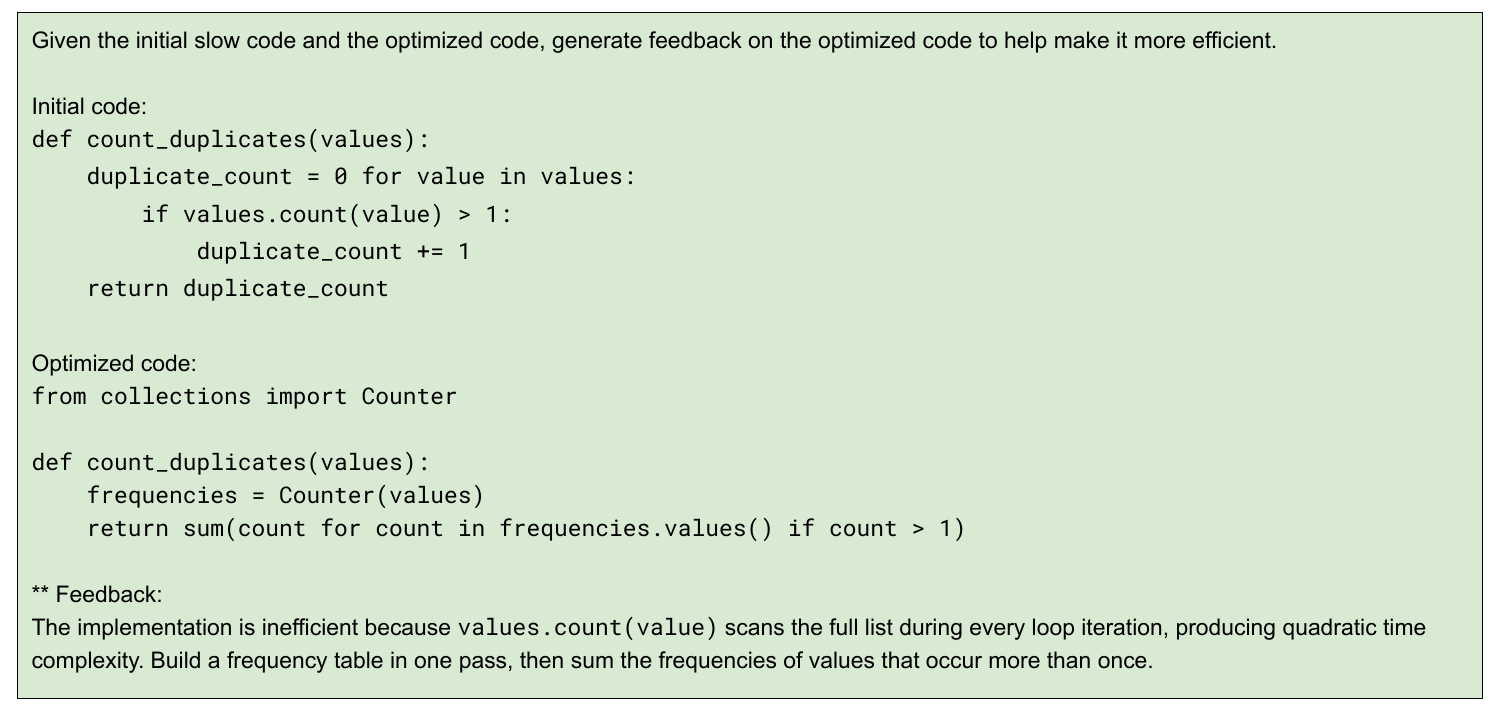}
    \caption{Critic prompt for PIE}
    \label{prompts_PIE_critic}
\end{figure*}

\begin{figure*}[!t]
    \centering
    \includegraphics[width=\textwidth]{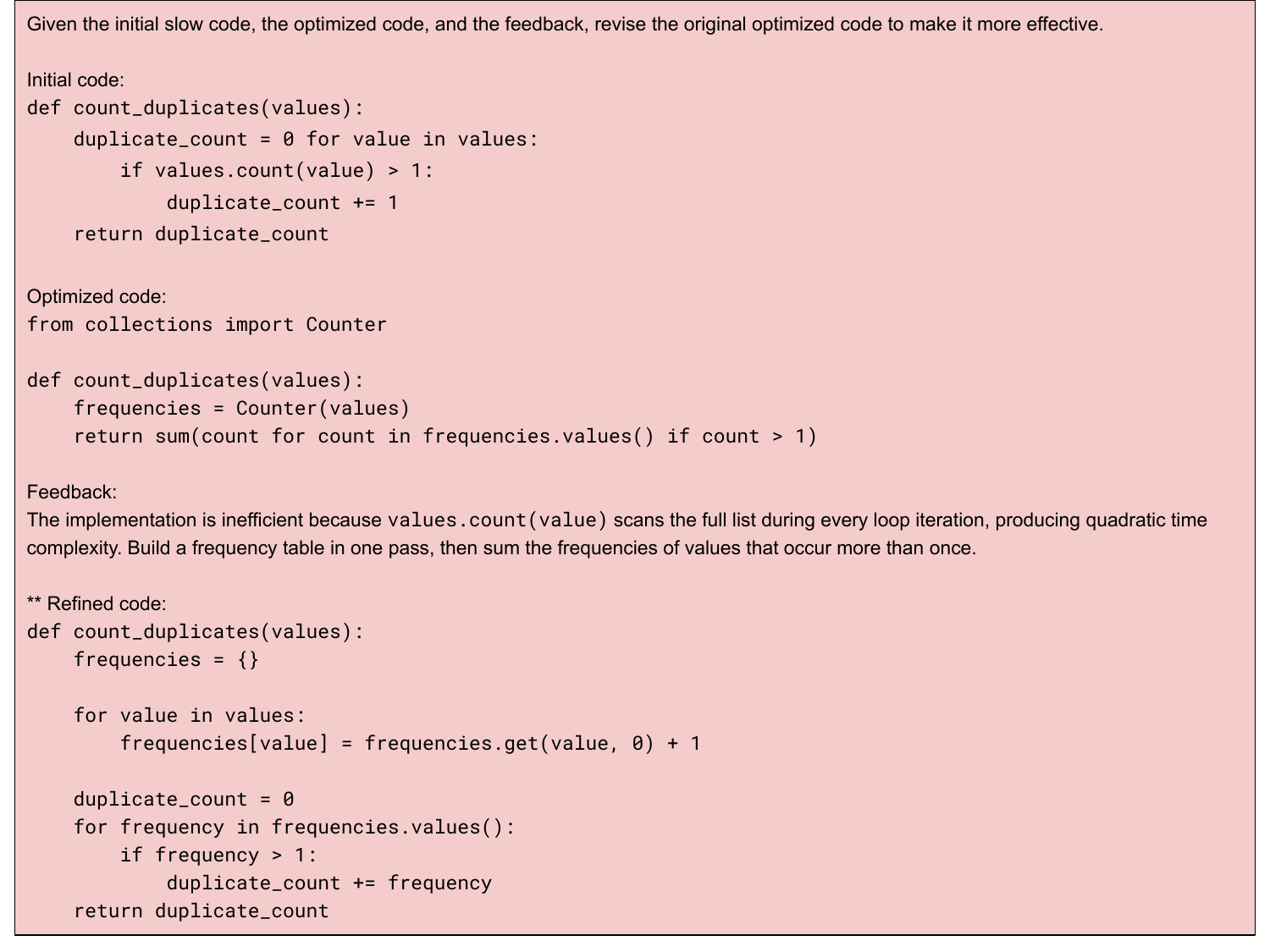}
    \caption{Refiner prompt for PIE}
    \label{prompts_PIE_refiner}
\end{figure*}

\begin{figure*}[!t]
    \centering
    \includegraphics[width=\textwidth]{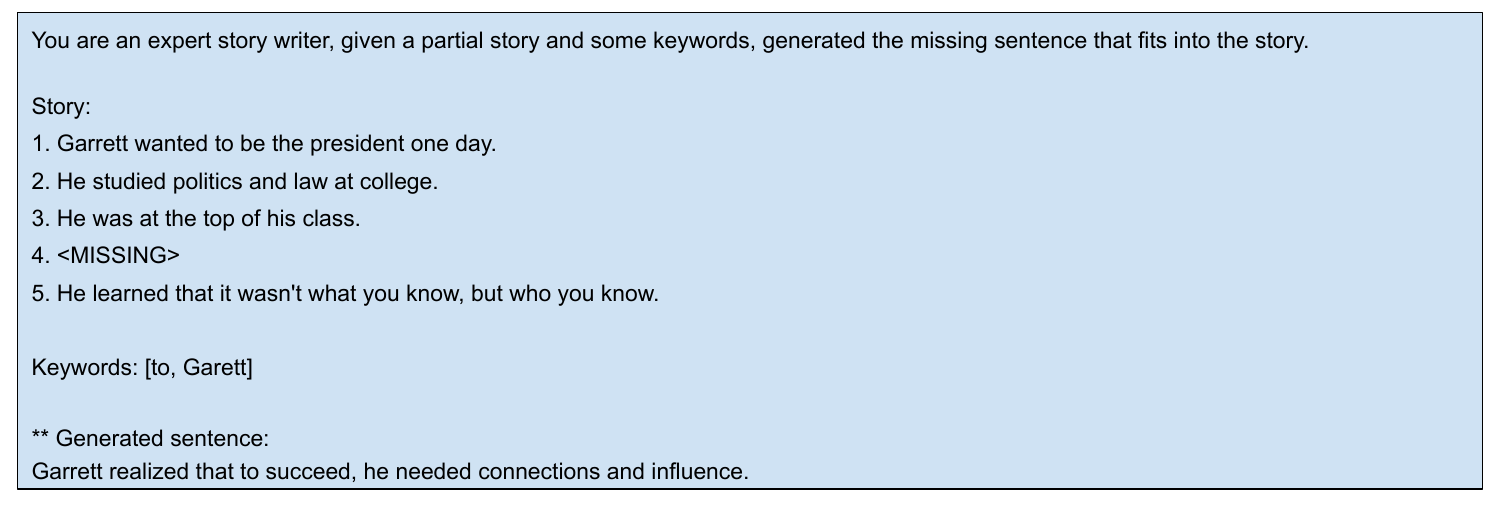}
    \caption{Generator prompt for CollaboSentGen}
    \label{prompts_collabosentgen_gen}
\end{figure*}

\begin{figure*}[!t]
    \centering
    \includegraphics[width=\textwidth]{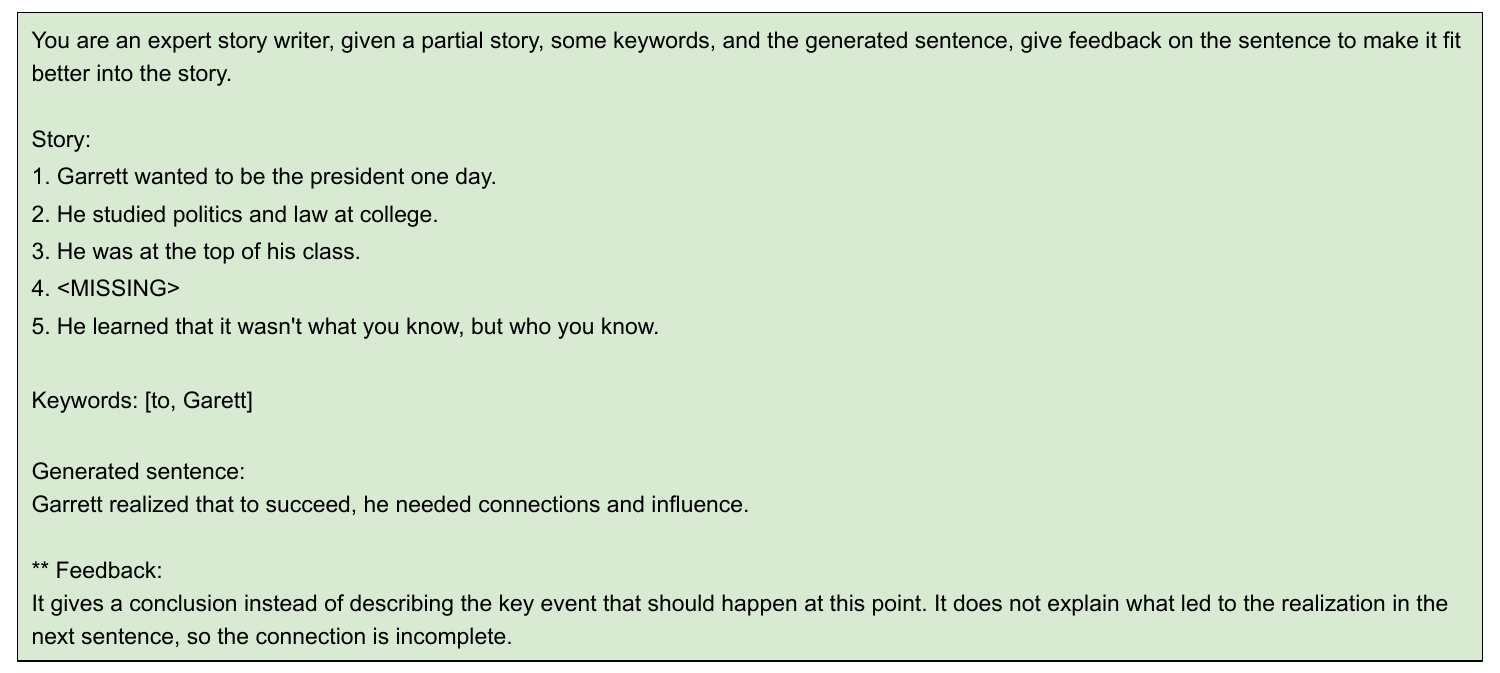}
    \caption{Critic prompt for CollaboSentGen}
    \label{prompts_collabosentgen_critic}
\end{figure*}

\begin{figure*}[!t]
    \centering
    \includegraphics[width=\textwidth]{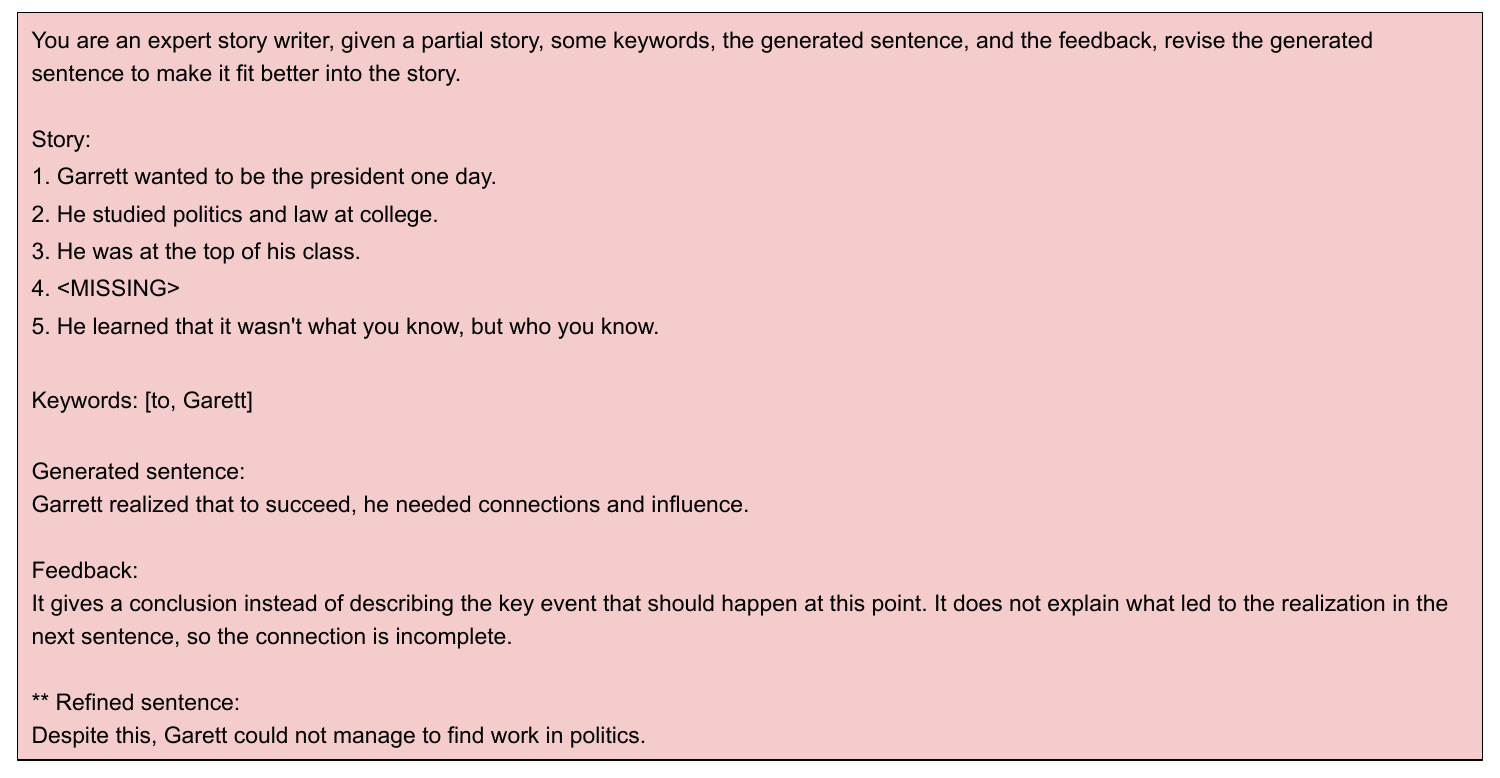}
    \caption{Refiner prompt for CollaboSentGen}
    \label{prompts_collabosentgen_refiner}
\end{figure*}

The prompts used in our experiments are organized as follows:

\begin{itemize}
    \item Meeting Plan: Figures~\ref{prompts_meeting_plan_gen}--\ref{prompts_meeting_plan_refiner}
    \item CNN/DailyMail: Figures~\ref{prompts_cnnDaily_gen}--\ref{prompts_cnnDaily_refiner}
    \item ZebraLogic: Figures~\ref{prompts_zebralogic_gen}--\ref{prompts_zebralogic_refiner}
    \item PIE: Figure~\ref{prompts_PIE_gen} -- \ref{prompts_PIE_refiner}
    \item CollaboSentGen: Figures~\ref{prompts_collabosentgen_gen}--\ref{prompts_collabosentgen_refiner}
\end{itemize}

\clearpage

\end{document}